\documentclass[a4paper,fleqn]{cas-sc}

\usepackage[numbers]{natbib}
\usepackage{amsmath}
\usepackage{url}            
\usepackage{booktabs}       
\usepackage{amsfonts}       
\usepackage{nicefrac}       
\usepackage{microtype}      
\usepackage{lipsum}
\usepackage{graphicx}
\usepackage{amsthm}
\usepackage{soul}
\usepackage[nameinlink,capitalise]{cleveref}
\usepackage{tikz}
\usepackage{tikz-3dplot}
\usepackage{subcaption}
\usepackage[dvipsnames]{xcolor}
\usepackage{comment}
\usepackage{bm}
\usepackage{relsize}
\usepackage{pgfplots}
\usepackage{ifthen}
\usepackage[most]{tcolorbox}
\usepackage{amssymb}
\usepackage{amsmath}
\usepackage{rotating}
\usetikzlibrary{arrows,shapes,snakes, decorations.pathmorphing,backgrounds,positioning,arrows.meta,intersections,decorations.pathreplacing,shapes.misc,angles, quotes,perspective}
\usetikzlibrary{positioning,fit}
\usepackage{tikz}
\usepackage{amsmath}

\usepackage{lscape}

\usetikzlibrary{decorations.pathmorphing, positioning}

\tikzstyle{mybox} = [text=black, very thick,
    rectangle, rounded corners, inner sep=10pt, inner ysep=20pt]
\tikzstyle{fancytitle} =[text=black]

\newcommand{\yslant}{0.5}
\newcommand{\xslant}{-0.6}

\newcommand\simplecuboid[4]{%
  \fill[#4!45!white] (tpp cs:x=0,y=0,z=#3)
    -- (tpp cs:x=0,y=#2,z=#3)
    -- (tpp cs:x=#1,y=#2,z=#3)
    -- (tpp cs:x=#1,y=0,z=#3) -- cycle;
  \fill[#4!70!white]  (tpp cs:x=0,y=0,z=0)
    -- (tpp cs:x=0,y=0,z=#3)
    -- (tpp cs:x=0,y=#2,z=#3)
    -- (tpp cs:x=0,y=#2,z=0) -- cycle;
  \fill[#4!25!white] (tpp cs:x=0,y=0,z=0)
    -- (tpp cs:x=0,y=0,z=#3)
    -- (tpp cs:x=#1,y=0,z=#3)
    -- (tpp cs:x=#1,y=0,z=0) -- cycle;}

\crefname{section}{Sec.}{Sec.}
\crefname{Line}{line}{§§}
\crefname{figure}{Fig.}{Fig.}
\crefname{table}{Tab.}{Tab.}
\crefname{algorithm}{Alg.}{Alg.}
\crefname{appendix}{Appx.}{§§}
\crefname{definition}{Def.}{Def.}
\crefname{equation}{Eq.}{Eq.}
\crefname{remark}{Remark.}{Remark.}
\crefname{theorem}{Theorem.}{Theorem.}

\newtheorem{definition}{Definition}

\newtheorem{remark}{Remark}
\newtheorem{theorem}{Theorem}

\newcommand{\x}[1]{\textcolor{orange}{#1}}

\begin{document}
\let\WriteBookmarks\relax
\def\floatpagepagefraction{1}
\def\textpagefraction{.001}
\shorttitle{Geometry is All You Need: A Unified Taxonomy of Matrix and Tensor Factorization for Compression of Generative Language Models}
\shortauthors{M. Xu et~al.}

\title [mode = title]{Geometry is All You Need: A Unified Taxonomy of Matrix and Tensor Factorization for Compression of Generative Language Models }                      
\author[1]{Mingxue Xu}[type=editor,
                        orcid=0000-0002-9942-0034]
\ead{m.xu21@imperial.ac.uk}
\ead[url]{https://mingxue-xu.github.io/}

\author[1]{Sadia Sharmin}[type=editor]
\ead{s.sharmin21@imperial.ac.uk}

\author[1]{Danilo P. Mandic}[type=editor, orcid=0000-0001-8432-3963]
\ead{d.mandic@imperial.ac.uk}
\ead[url]{https://www.commsp.ee.ic.ac.uk/~mandic/}

\affiliation[1]{organization={Department of Electrical and Electronic Engineering, Imperial College London},
                addressline={Exhibition Rd, South Kensington}, 
                city={London},
                postcode={SW7 2AZ}, 
                country={United Kingdom}}

\begin{abstract}
Matrix and tensor-guided parametrization for Natural Language Processing (NLP) models is fundamentally useful for the improvement of the model's systematic efficiency, as proved by recent research progress~\cite{hu2021lora,Dao2022MonarchES,zhao2024galore}.
However, the internal links between these two algebra structures and language model parametrization are poorly understood. 
Also, the existing matrix and tensor research is math-heavy and far away from machine learning (ML) and NLP research concepts. These two issues result in the recent progress on matrices and tensors for model parametrization being more like a loose collection of separate components from matrix/tensor and NLP studies, rather than a well-structured unified approach, further hindering algorithm design. 
To this end, we propose a unified taxonomy, which bridges the {\it matrix/tensor compression approaches} and {\it model compression concepts in ML and NLP research}. 
Namely, we adopt an elementary concept in linear algebra, that of a {\bf subspace}, which is also the core concept in geometric algebra, to reformulate the matrix/tensor and ML/NLP concepts (e.g. attention mechanism) under one umbrella.
In this way, based on our subspace formalization, typical matrix and tensor decomposition algorithms can be interpreted as {\bf geometric transformations}. 
Finally we revisit recent literature on matrix- or tensor-guided language model compression, rephrase and compare their core ideas, and then point out the current research gap and potential solutions.  

\end{abstract}

\begin{keywords}
Language Processing \sep Model Compression \sep Low-rank Factorization \sep Linear Algebra \sep Geometric Transformation
\end{keywords}

\maketitle
\section{Introduction}
\begin{figure}[h!]
\centering
\begin{tikzpicture}
\usetikzlibrary{positioning,fit}

    \node[rectangle, rounded corners=5, minimum width=35em, minimum height=4em, draw=none, fill=red!20,fill opacity=0.5] (target) at (0, 0) {};
    \node[rectangle, rounded corners=5, minimum width=35em, minimum height=4em, draw=none, fill=cyan!50,fill opacity=0.3] (tech) at (15em, 0) {};
    \node[text=red,font=\large] at ($(target.center)+(-12em,+1)$) {Application};
    \node[text=blue,font=\large] at ($(tech.center)+(+11em,+1)$) {Methodology};
    \node[text=violet,font=\large] (overlap) at ($(target.center)+(+7.5em,+1)$) {Literature};

    \node[rectangle, rounded corners=15, minimum width=12em, minimum height=3em, very thick, draw=orange, fill=yellow!70,fill opacity=0.3] (vec) at ($(overlap.center)+(0,-4.5)$) {};
    \node[font=\large,text=black] at ($(vec.center)$) { Geometric Algebra};
    \node[text=orange,font=\large] at ($(vec.center)+(-0.9,+0.75)$) {Solution};
    
    \draw[-stealth, very thick] (vec.west) to[bend left] ($(target.south)+(-3,0)$);
    \draw[-stealth, very thick] (vec.east) to[bend right] ($(tech.south)+(+3,0)$);
    \draw[-stealth, very thick] (vec.north) -- ($(overlap.center)+(0,-5em)$);

    \node[rectangle, rounded corners=15, minimum width=12em, minimum height=3em, draw, very thick,fill=white] (reform) at ($(vec.center)+(-5,1.6)$) {};
    \node[color=black] at ($(reform.center)$) {reformulate (\cref{sec:application})};
    
    \node[rectangle, rounded corners=15, minimum width=12em, minimum height=3em,draw, very thick,fill=white] (unify) at ($(vec.center)+(+5,1.6)$) {};
    \node[color=black] at ($(unify.center)$) {unify (\cref{sec:decompose})};

    \node[rectangle, rounded corners=15, minimum width=15em, minimum height=3em, draw, very thick,fill=white] (revisit) at ($(vec.center)+(0,+1.6)$) {};
    \node[color=black] at ($(revisit.center)$) {interpret and compare  (\cref{sec:literature})};



    \node[] at ($(overlap.center)+(0,-0.75)$) {\cite{Dao2022MonarchES,edalati2022kronecker,tahaei2021kroneckerbert,chen2018groupreduce,chen2021drone,Lan2019ALBERTAL,ben-noach-goldberg-2020-compressing,li2023losparse}};
    \node[] at ($(overlap.center)+(0,-1.25)$) {\cite{wang2022exploring,chekalina2023efficient,hrinchuk2019tensorize,liu2021enabling,NEURIPS2019_dc960c46,li-etal-2022-hypoformer,qiucompute}};
    
    \node[color=black] at ($(target.center)+(-10em,0.25)$) {Natural Language Processing};
    \node[color=black] at ($(target.center)+(-10em,-0.25)$) {Neural Network Compression};

    \node[color=black] at ($(tech.center)+(+10em,+0.25)$) {Matrix Factorization};
    \node[color=black] at ($(tech.center)+(+10em,-0.25)$) {Tensor Factorization};

    \draw[black!50, thick, dashed] ($(tech.center)+(-16em,+0.5)$) rectangle ($(tech.center)+(+16em,0.05)$);
    \draw[black!50, thick, dashed] ($(tech.center)+(-16em,-0.05)$) rectangle ($(tech.center)+(+16em,-0.5)$);

\end{tikzpicture}
\caption{Addressed issues, solutions and paper structure overview.} 
\label{fig:outline}
\end{figure}
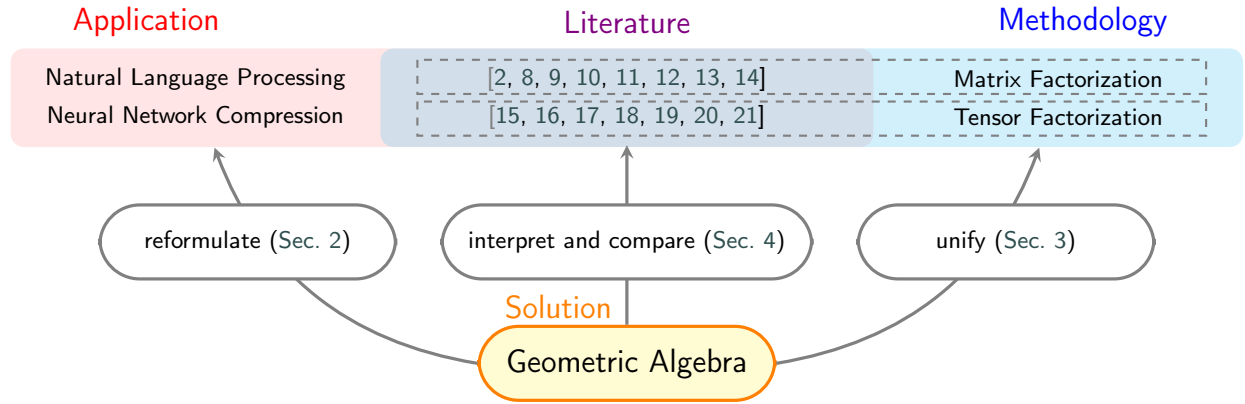

Matrix and tensor factorization for generative language model compression refers to approximating weight matrices with a product of smaller matrices or tensors (higher-dimensional matrix) to reduce the model size, thereby alleviating memory usage or computation overhead while maintaining language task performance.
Since the emergence of successful solutions regarding neural network compression (see~\cref{fig:outline}), such a factorization approach has attracted massive interest.
Compared with other model compression approaches like knowledge distillation and quantization, matrix and tensor factorizations have unique advantages: 1) they require less training effort than that based on knowledge distillation; 2) they can be hardware-independent, where quantization cannot. 

However, factorization for generative language model compression is poorly defined and understood. In the current literature, the application side (i.e. language tasks and compression) and methodology side (matrix and tensor factorization) are separate, and the potential interaction between the algebra structure and deep learning research is ignored. This includes model capacity, that is, how much information deep neural networks can memorize~\citep{AllenZhu2024PhysicsOL}, and model expressivity, that is, what kind of functions deep neural networks can deal with~\cite{montufar2014number,pmlr-v70-raghu17a}. 
Furthermore, there is also an inappropriate separation between matrix-based solutions and tensor-based solutions (i.e. two separate reference lines in the central literature part of~\cref{fig:outline}), though mathematically matrix algebra represents tensor algebra for order-2 tensors. This results in the difficulties of comparing the compression algorithms and the empirical results of the two, to identify the current research gap. 
Both these two isolated issues discard the possibilities of matrix or tensor factorization promoting deep learning research with its algebraic properties. 
The third dilemma comes from matrix and tensor research. The terminologies of matrix algebra are not always consistent (e.g. the definition of ``direct sum''~\citep[Section 2.1]{gentle2023matrix}). The terminologies for tensors, on the other hand, are hard to understand since they come from various scientific fields (e.g. psychometrics, physics and quantum chemistry). These inconsistent and unfriendly terminologies block the model compression community from using these algebra structures to the full for further practical impacts.

To address the three above issues, this paper proposes an elementary and unified terminology of matrix and tensor factorization for generative language model compression. Specifically, we utilize a concept undergraduate-level linear algebra (also the core concept in geometric algebra), that of a {\it subspace}, to unify the concepts in tensor- and matrix-based approaches, as shown in~\cref{fig:outline}. It is our hope that in this way these methodologies can become more approachable for people with broader science, technology, engineering, and mathematics backgrounds. 
We also formalize the concepts in generative language model compression in the context of ``subspace'', e.g. parameter space, neural network compression and language modelling. 
We then give the geometric interpretation of these reformulated concepts and common matrix/tensor compression methodologies (e.g. SVD, Kronecker product and tensor decomposition), smoothly bridging compression methodologies and deep learning concepts, while aiding the understanding of tensor operations. Based on the proposed taxonomy, in~\cref{sec:literature}, we list the recent literature, interpret and compare their approaches and results, and reveal the current research gaps together with suggesting further directions.

The structure of this paper is shown in the middle panel of~\cref{fig:outline}. We introduce our formalization preliminaries and the reformulated concepts of generative language model compression in a geometric way in~\cref{sec:application}. We then provide the interpretation of matrix and tensor factorization methodologies geometrically in~\cref{sec:decompose}. Subsequently, we put everything in~\cref{sec:application,sec:decompose} together, interpret the existing relevant work under our terminology framework, compare them and point out the current research gap in~\cref{sec:literature}. Finally, we summarise our contribution, limitations and future work in~\cref{sec:conclusion}.

\section{Geometric Algebra for Model Compression}\label{sec:application}
In this section, we reformulate the concepts and problems regarding deep neural network (DNN) compression and natural language processing (NLP, probabilistic language modelling as a case study), in order to ``compile'' the matrix and tensor factorization under the same framework as generative language model compression - a framework built upon geometric algebra.

\subsection{Parameter Space and Model Compression}\label{sec:params}

A typical parametrization of neural networks~\citep[Section 3.3]{choromanska2015loss}, which we will follow in our proposed taxonomy is as follows. Let ${W}=\{{\bf W}_1,{\bf W}_2,\ldots, {\bf W}_l\}$ be the set of all weight matrices of a deep neural network model $f$, where $l$ is the number of neural network layers.
Denote by $\mathbb{R}$ the set of real numbers, and $n$ the parameter count of the model $f$, so that we have the upper bound of model parameter count $\sup(f)=\sum^{l}_{i=1} \vert {\bf W}_i \vert$, $i=1,2,\ldots,l$, where $\vert {\bf W}_i \vert$ stands for the number of elements of matrix ${\bf W}_i$ and $\sup(f)$ is the upper bound parameter count of deep neural network model $f$. 

As a step forward from this parametrization, we decrease the layers (i.e. layer number) and dimensionality (i.e. the concept ``matrices'') of the set ${W}$, flatten the elements of all the weight matrices into a vector ${\bf v}$, with a comma-delimited form ${\textbf{v}}=(v_1,v_2,\ldots, v_n)$. We can also represent ${\bf v}$ with $\sup(f)$-dimensional unit vectors ${\textbf e}_1=(1,0,0,\ldots, 0)$, ${\textbf e}_2=(0,1,0,\ldots, 0)$, \ldots, ${\textbf e}_n=(0,0,0,\ldots, 1)$, like ${\bf v} = v_1 {\bf e}_1 + v_2 {\bf e}_2 + \cdots + v_n {\bf e}_n$. Here, $\{{\textbf e}_1, {\textbf e}_2,\ldots,{\textbf e}_n\}$ is a basis set of the vector space which contains ${\bf v}$, while $\{v_1, v_2,\ldots,v_n\}$ is the set of real value coefficients of vector ${\bf v}$. 

Denote by $V$ a set that contains all possible linear combinations of $\{{\textbf e}_1, {\textbf e}_2,\ldots,{\textbf e}_n\}$ and $ \{v_1, v_2,\ldots,v_n\}$, and a vector space $\mathbb{V}$ which is an algebra structure that contains vector set $V$ and scalar multiplication and vector addition on set $V$. For simplicity, we use $\mathbb{V} \subseteq \mathbb{R}^{n}$ to denote a vector space that contains all vectors in $\mathbb{R}^{n}$, where $\mathbb{R}$ is a set of real numbers. 
We shall now give the following definition of the model parameter vector and parameter space. 

\begin{definition}[Parameter Vector and Parameter Space]\label{def:vec} Given a vector space $\mathbb{P} \subseteq \mathbb{R}^{n}$, a vector ${\bf v} \in \mathbb{P}$ is the parameter vector with a coefficient set $\{v_i \vert i=1,\ldots,n\}$. Then, $\mathbb{P}$ is called the parameter space, and $s({\bf v}) = \sum^{n}_{i=1} \left ( \text{sgn} (v_i)\right )^2$ is the parameter count of ${\bf v}$, where $\text{sgn}$ is the sign function.  
\end{definition}

Note that $s({\bf v})$ is the number of the non-zero elements in $W$, $s({\bf v})\leq \sup(f)$. If $s({\bf v})$ is significantly smaller than $\sup(f)$, we say the weight matrices of model $f$ are sparse. A model compression technique called pruning, which reduces the size of a neural network by removing less important parameters (i.e. zeros or nearly zero) or weights, can be seen as removing the subvectors with zero coefficients in $\mathbb{V}$. However, a systematic implementation of this, like only saving the indices of the less important parameters is outside the scope of this paper.

Suppose there is another subset of $V$, denoted as $\tilde{V}$. The vector space built up on $\tilde{V}$ is called the subspace of $\mathbb{V}$, and denoted as $\tilde{\mathbb{V}}$. Together with~\cref{def:vec}, this now gives the following formal definition of deep neural network compression. 

\begin{definition}[Low-rank Neural Network Compression]\label{def:mc} Let $\tilde{\mathbb{P}}$ be a subspace of $\mathbb{P}$, and $\exists \tilde{\bf v} \in \tilde{\mathbb{P}}$, $s(\tilde{{\bf v}}) < s({\bf v})$. The mapping $g: \mathbb{P} \rightarrow \tilde{\mathbb{P}}$ for extracting such a subspace $\tilde{\mathbb{P}}$ is called low-rank neural network compression, where $\texttt{CR}({\bf v}, \tilde{\bf v}) = \frac{s({\bf v})-s(\tilde{\bf v})}{s(\tilde{\bf v})}$ is the compression ratio function, while a compression instance with a given threshold $\eta$, which satisfies $\texttt{CR}({\bf v}, \tilde{\bf v})>\eta$, is called $\eta$-compression.
\end{definition}

The formalization in~\cref{def:mc} is rather general. Taking the notion of a {\it subspace} in a broad sense, the compression output -  smaller matrices/tensors can be represented with the same unit vectors as the original weight matrices, and the shrunk parameters can be seen as the corresponding coefficients of these changes to $0$.  
For example, the most straightforward approach for~\cref{def:mc} would be to break the original weight matrices into smaller matrices with SVD, Kronecker product~\cite{edalati2022kronecker,tahaei2021kroneckerbert} or block-diagonal matrices~\cite{Dao2022MonarchES,Dao2020KaleidoscopeAE,Dao2019LearningFA}

\subsection{Generative Language Modelling with Parameter Vector}

For a more general language task description rather than case-by-case, we focus on the modelling process from~\cite{radford2019language}, which is based on language modelling task and takes other language tasks as downstream tasks. For the more complex language tasks like reasoning in~\cref{sec:llm}, we will give a higher-level geometric interpretation with current popular large language models. 

\begin{definition}[Generative Language Modelling]\label{def:lm} Given a sequence of tokens $(x_1, x_2, \ldots, x_t)$ of length $t$, the goal of generative language modelling is to find a factorized joint probabilities $ p(x_t) = \prod^{l}_{i=1}p(x_t \vert x_1, \ldots, x_{t-1}) $. Denote the token sequence $(x_1, x_2, \ldots, x_t)$ as a vector ${\bf x}$, the probability of token sequence ${\bf x}_i$ of length $i$ is
\begin{equation}\label{eq:lm}
    y=p({\bf x}) = \prod^{t}_{i=1} p({\bf x}_i \vert {\bf x}_{i-1}),\quad {\bf x}_i = (x_1, \ldots, x_i, 0, \ldots, 0).
\end{equation}
\end{definition}

For a given neural network model $f$ and its parameter vector ${\bf v}$ defined in ~\cref{def:vec}, denote by $\mathbb{X}$ the input space of $f$, which contains the token sequences. Denote by $\mathbb{Y}$ the output probability space of $f$ and $\forall y \in \mathbb{Y}$, $y \in [0,1]$. 
Given some loss function $\mathcal{L}$, we can obtain the loss of certain model prediction, $\hat{\bf y}=f({\bf x})$, and the ground truth label, ${\bf y}$, as $\mathcal{L}(\hat{\bf y},{\bf y})$. In this way, the learning objective for \cref{def:lm} is $\min \mathcal{L}({\bf y},\hat{\bf y})$. 

\begin{figure}[h!]
\centering

\begin{subfigure}{\textwidth}
\input{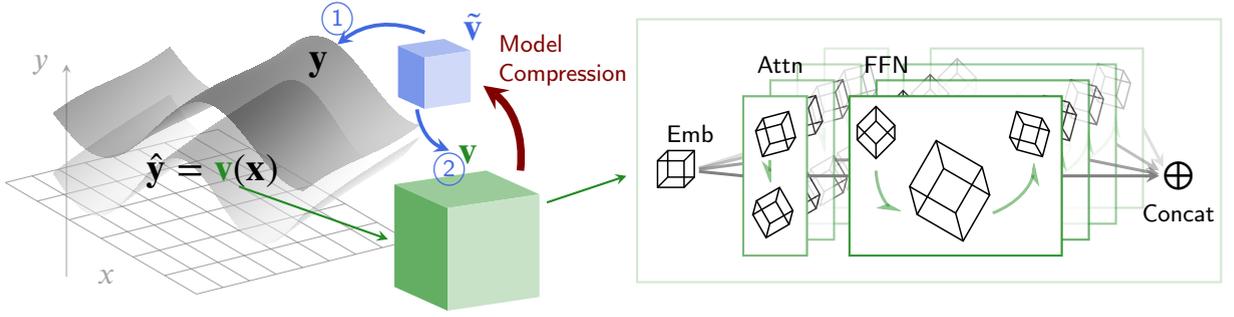}
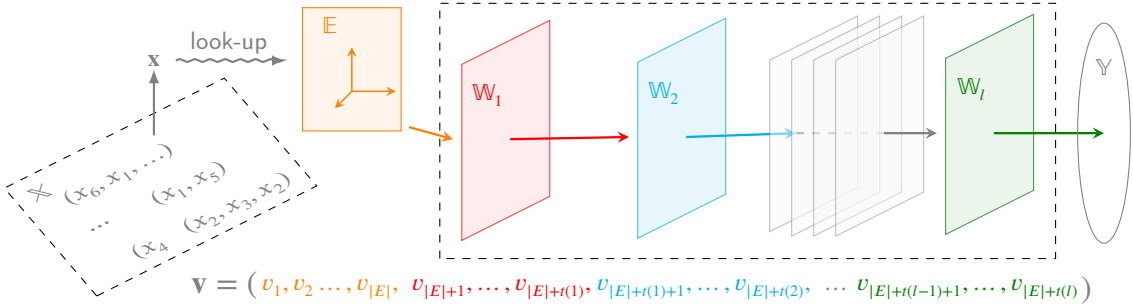
\caption{Model compression (defined in~\cref{def:lmc}) in the context of the parameter vector, with a transformer structure as a case study. On the left side, the two optimization goals for model compression are ${\large{\textcircled{\small 1}}}$ accuracy (${\hat{\bf y}} \rightarrow {\bf y}$ or optimize $\texttt{ACC}({\bf{y}}, \hat{\bf{y}})$ in~\cref{def:acc}), and ${\large{\textcircled{\small 2}}}$ fidelity (${\bf v} \rightarrow \tilde{\bf v}$ in~\cref{def:fdl}). The right side is the transformer structure discussed in~\cref{sec:lm-module}.}
\label{fig:lm-mc} 
\end{subfigure}
\hfill
\begin{subfigure}{\textwidth}
\begin{tikzpicture}[scale=0.58,every node/.style={minimum size=1cm},on grid]

	
	\begin{scope}[
		yshift=0,
		every node/.append style={yslant=\yslant,xslant=\xslant},
		yslant=\yslant,xslant=\xslant
	]
		\draw[black, dashed, thin] (0,0.5) rectangle (5,4); 
            \node at (0.5,3.5) {$\mathbb{X}$};
            \node at (1.5,1) {$(x_4$};
            \node at (3,2) {$(x_1, x_5)$};
            \node at (3.5,1) {$(x_2, x_3, x_2)$};
            \node at (2,3) {$(x_6, x_1, \ldots)$};
            \node at (1,2) {$\cdots$};
	\end{scope} 
        \draw[thick, -latex, decoration={segment length=2mm, amplitude=0.2mm}, decorate] (1,4) to (1,5.5);
        \node at (1,5.7) {${\bf x}$};

        \begin{scope}

        \node[rectangle, thin,fill opacity=0.75,fill=orange!10, minimum width=4em, minimum height=5em,draw=orange] (E) at (5.5,5.5) {};

        \draw[line width=0.2mm, -stealth,orange] (5.5,5,0) -- (6.5,5,0) node[anchor=north east]{};
    \draw[line width=0.2mm,-stealth,orange] (5.5,5,0) -- (5.5,6,0) node[anchor=north west]{};
    \draw[line width=0.2mm,-stealth,orange] (5.5,5,0) -- (5.5,5,1) node[anchor=east]{};
    
    \node[orange] at (5,6.5) {$\mathbb{E}$};
        
        \end{scope}

        \draw[snake,thick, -latex, decoration={segment length=2mm, amplitude=0.2mm}, decorate] (1.5,5.7) to (4,5.7);
        \node at (2.7,6.1) {look-up};

        \draw[black, dashed, thin] (7.5,1.2) rectangle (21.5,7); 
    \begin{scope}[
		yshift=-13.5em,
		yslant=1-\yslant
	]
    
    \draw[red, thin, fill=red!10,opacity=.75] (8,2) rectangle (10,6);
    
    \node[red] (W_1) at (8.6,5) {$\mathbb{W}_1$};
    \end{scope}
    
    \begin{scope}[
		yshift=-19.5em,
		yslant=1-\yslant
	]
        
    \draw[cyan, thin, fill=cyan!10,opacity=.75] (12,2) rectangle (14,6);
    \node[cyan] (W_2) at (12.6,5) {$\mathbb{W}_2$};
    \end{scope}

    \begin{scope}[
		yshift=-30em,
		yslant=1-\yslant
	]
        
    \draw[Green, thin, fill=Green!10,opacity=.75] (19,2) rectangle (21,6);
    \node[Green](W_l) at (19.6,5) {$\mathbb{W}_l$};
    \end{scope}

    \draw[line width=0.3mm,-stealth,orange] ($(E.south)+(1.3,0.1)$) --($(W_1)+(-0.8,-1)$);
    \draw[line width=0.3mm,-stealth,red] ($(W_1)+(0.5,-1)$) -- ($(W_2)+(-0.8,-1)$);
    \draw[line width=0.3mm,-stealth,cyan] ($(W_2)+(0.5,-1)$) -- ($(W_l)+(-4,-1)$);

    \begin{scope}[
		yshift=-25.5em,
		yslant=1-\yslant
	]
    \draw[gray, thin, fill=gray!5,opacity=.5] (15,2.6) rectangle (17,6.5);
    \draw[line width=0.3mm,dashed] ($(W_l)+(-4,1)$) -- ($(W_l)+(-2,0)$);
        \draw[gray, thin, fill=gray!5,opacity=.5] (15.5,2.25) rectangle (17.5,6.25);
    
    \draw[gray, thin, fill=gray!5,opacity=.5] (16,2) rectangle (18,6);
    \draw[gray, thin, fill=gray!5,opacity=.5] (16.5,1.75) rectangle (18.5,5.75);
    
    \end{scope}

    \draw[line width=0.3mm, gray,-stealth] ($(W_l)+(-2,-1)$) -- ($(W_l)+(-0.8,-1)$);

    \draw[line width=0.3mm,-stealth,Green] ($(W_l)+(0.5,-1)$) -- ($(W_l)+(3,-1)$) node[anchor=east]{};
    \node[gray] at ($(W_l)+(3,+0.5)$) {$\mathbb{Y}$};

    \node[font=\large] at ($(E.center)+(-2.9,-4.9)$) {${\bf v}=($};
    \node[orange] at ($(E.center)+(-0.5,-5)$) {$v_1, v_2\ldots, v_{\vert E \vert},$};

    \node[red] at ($(E.center)+(3.4,-5)$) {$v_{\vert E\vert+1},\ldots,v_{\vert E \vert+ t(1)},$ };

    \node[cyan] at ($(E.center)+(8,-5)$) {$v_{\vert E \vert+ t(1)+1},\ldots,v_{\vert E \vert+ t(2)},$ };

    \node[] at ($(E.center)+(11,-5)$) {$\ldots$ };
    \node[Green] at ($(E.center)+(14,-5)$) {$v_{\vert E\vert+t(l-1)+1},\ldots,v_{\vert E \vert+ t(l)}$ };
    \node[font=\large] at ($(E.center)+(16.7,-5)$) {$)$ };

    \draw ($(W_l)+(3,-1)$) ellipse (0.6cm and 2.5cm);
       
\end{tikzpicture}
\caption{Generative language modelling with a layerwise parameter vector; $t(\cdot)$ is the index indicator function of ${\bf v}$, $t(i)=\sum^{i}_{k=1} \vert {\bf W}_k \vert$.} 
\label{fig:lm}
\end{subfigure}

\caption{Parametrization of generative neural networks in a vector space, and the process of model compression based on, taking a typical transformer~\cite{vaswani2017attention} forwarding pass as case study.}
\end{figure}

Informally, we can also represent the prediction as $\hat{\bf y} = {\bf v}({\bf x})$, here ${\bf v}(\cdot)$ is a series of transformations defined by the vector ${\bf v}$. A vector space version of neural probabilistic language modelling~\cite{nplm} is shown in~\cref{fig:lm} and on the left side of~\cref{fig:lm-mc}.

\subsection{Generative Language Model Compression}\label{sec:lmc}

Model compression is somewhat similar to data compression, that is, using fewer data/parameters to achieve similar performance or representation as the original ones. However, when the objective changes from data to a neural network model, the goals and evaluation metrics change. For data compression, the goal is to reduce the size of the original data, but retain the necessary information that can be effectively restored to the original data. However, for model compression, except restoring to the original model weights, we care more about the task accuracy of the neural network model that can be reserved. Thus, when talking about language model compression, our concerns from both data compression and model performance perspectives are as follows: 
\begin{enumerate}
    \item {\bf Accuracy}: To what extent, after adopting the smaller matrices/tensors into the neural network models, the model can have a similar performance (e.g. modelling the data distribution) to that of the original model;
    \item {\bf Fidelity}: The extent to which the smaller matrices/tensors can be recovered to the original ones?
\end{enumerate}

This leads us to the formal definition of these two model compression optimization goals: fidelity and accuracy.

\begin{definition}[Accuracy]\label{def:acc} Given a loss function, $\mathcal{L}$ (typically returns non-negative values), an input, ${\bf x}$, a label, ${\bf y}$ and a parameter vector, ${\bf v}$, the accuracy of low-rank language model compression is given by
\begin{equation}
    \texttt{ACC}({\bf{y}}, \hat{\bf{y}}) = \frac{\sum^{l}_{i=1}\mathcal{L}({\bf{y}}, \hat{\bf{y}}) }{l}, \quad \hat{\bf y} = {\bf v}({\bf x})
\end{equation} 
\end{definition}

\begin{definition}[Fidelity]\label{def:fdl} Assume that a low-rank model compression mapping, $g: \mathbb{P} \rightarrow \tilde{\mathbb{P}}$, has an inverse, $g^{-1}: \tilde{\mathbb{P}},  \rightarrow {\mathbb{P}}$. Then, the fidelity function $\texttt{FDL}(\cdot)$ is defined as the Mean Absolute Error (MAE) of the original parameter vector ${\bf v}$ and the reconstructed parameter vector $\hat{\bf v}$, and has the form 
\begin{equation}
    \texttt{FDL}({\bf v}, \hat{\bf v}) = \frac{\sum^{n}_{i=1}\vert v_i - \hat{v}_i  \vert}{n}, \quad\hat{\bf v}=g^{-1}(\tilde{{\bf v}}).
\end{equation} 
\end{definition}

By considering~\cref{def:vec,def:mc,def:lm,def:fdl,def:acc} together, we obtain an integrated formalization of generative language model compression as follows.
\begin{definition}[$(\delta_1, \delta_2, \eta)$ -  Language Model Compression]\label{def:lmc} Consider an input domain, $\mathbb{X}$, output domain, $\mathbb{Y}$, and a loss function, $\mathcal{L}$, for language modelling. For a dataset $D = \{({\bf x}, {\bf y}) \vert {\bf x}\in \mathbb{X}, {\bf y}\in \mathbb{Y} \}$, if the following conditions are satisfied, such a compression scheme is termed as a \ul{ $(\delta_1, \delta_2, \eta)$ -  Generative Language Model Compression}:
\begin{enumerate}
\item $\forall ({\bf x}, {\bf y}) \in D$, $\texttt{FDL}({\bf v}, \hat{\bf v})<\delta_1 \bigwedge \texttt{ACC}({\bf{y}}, \hat{\bf{y}})<\delta_2$, where $\hat{\bf y} = {\bf v}({\bf x})$;
\item for $g: {\bf v} \rightarrow \tilde{\bf v}$, compression ratio $\texttt{CR}({\bf v}, \tilde{\bf v})> \eta$.
\end{enumerate}   
\end{definition}

From~\cref{def:lmc}, if we only focus on the accuracy of the compressed language model, $\delta_1 \rightarrow \infty$ (can be set as a large value in practice), the fidelity will be ignored during the optimization process. Similarly $\delta_2 \rightarrow \infty$ if only fidelity is required. 

~\cref{fig:lm-mc} provides an interpretation of~\cref{def:lmc}, taking language modelling task as an example. As addressed in~\cref{def:lm}, language modelling uses a model ${\bf v}\in \mathbb{P}$, $\mathbb{P}\subset \mathbb{R}^{n}$ to approximate the distribution of $\bf y$, that is ${\bf v}({\bf x}) \rightarrow {\bf y}$. Model compression aims to find a vector $\tilde{\bf v}$, which lies in a subsapce of $\mathbb{P}$, such that $ s(\tilde{\bf v}) <  s({\bf v})$. There are two criteria for the process ${\bf v} \rightarrow \tilde{\bf v}$: $\large{\textcircled{\small 1}}$ if $\tilde{\bf v}$ can well approximate ${\bf y}$, and $\large{\textcircled{\small 2}}$ is $\tilde{\bf v}$ can be well reconstructed to ${\bf v}$. Criterion $\large{\textcircled{\small 2}}$ is the same criteria as for data compression, criteria $\large{\textcircled{\small 1}}$, however, indicates {\it language modelling is compression}~\cite{deletang2024language}.

\subsection{Weight Matrices/Tensors Composed by Subspaces}\label{sec:subspaces}
\subsubsection{Layerwise Parameterization}

In~\cref{def:vec,def:mc} we considered all the parameters of a model in one parameter vector for easing the problem definition. However, the current literature typically considers the factorization of each layer, that is, factorize weights for each layer, which is shown in~\cref{fig:lm}.
Thus for the following sections, we discuss model compression layer by layer.

For a neural network model $f = f_1 \circ f_2 \circ \cdots \circ f_l$, where $l$ is the maximum layer index of model $f$, and $\circ$ denotes function composition, by representing this form with parameter vectors in the parameter space $\mathbb{P}$, we have
\begin{align}
{\bf y} & = f({\bf v}, {\bf x}) = f_l({\bf v}_l,\ldots,f_3({\bf v}_3,f_2({\bf v}_2,f_1({\bf v}_1,{\bf x})))\cdots), \\
{\bf v} & = \sum^{l}_{i=1} {\bf v}_i,\quad {\bf v}_1, {\bf v}_2, \ldots, {\bf v}_l \in \mathbb{P}.\label{eq:sum-v}
\end{align}

In~\cref{eq:sum-v}, the summation for ${\bf v}_i$ can also represent concatenation in neural networks, which we shall discuss in~\cref{sec:concat}. 
If ${\bf v}_i$ is expressed in a comma-delimited form, we can define a vector index function $\texttt{idx}({\bf v}, i) = \sum^{i}_{j=1} s({\bf v}_{i})$, and its increment $\Delta \texttt{idx}({\bf v}, i) = \texttt{idx}({\bf v}, i) - \texttt{idx}({\bf v}, i-1)$ , to have 
\begin{equation}\label{eq:v-idx}
    {\bf v}_i = (\overbrace{0,\ldots,0}^{\texttt{idx}({\bf v}, i-1)},\overbrace{ v_{\texttt{idx}({\bf v}, i-1)+1}, v_{\texttt{idx}({\bf v}, i-1)+2}, \ldots,v_{\texttt{idx}({\bf v}, i)}}^{\Delta \texttt{idx}({\bf v}, i)}, \overbrace{ 0,\ldots, 0}^{\texttt{idx}({\bf v}, l) - \texttt{idx}({\bf v}, i)}).
\end{equation}

\subsubsection{Matrix and Tensor Composition}

Now that we have ${\bf v}_i$ as the parameter vector for the weight matrix of a concerned layer $i$, to form it into a matrix, denote by a $\mathbb{P}^{(i)}$ subspace whose basis set is same as $s({\bf v}_i)$, and two other subspaces $\mathbb{P}^{(i)}_{1}, \mathbb{P}^{(i)}_{2} \subseteq \mathbb{P}^{(i)}$, $\mathbb{P}^{(i)}_{1} \oplus \mathbb{P}^{(i)}_{2} = \mathbb{P}^{(i)}$, where
$\oplus$ denotes direct sum, which means the union of basis sets of two subspaces and the formation of a new vector space on this basis set union. 
Recall ${\bf v}_i$ is originally a weight matrix of layer $i$, ${\bf v}_i$ can then be considered as a vector in $\mathbb{P}$ which is composed by  $\mathbb{P}^{(i)}_{1}$ and $\mathbb{P}^{(i)}_{2}$. 

Then, the $(j,k)$th element of the matrix represented by ${\bf v}_i$ is $v_{j}v_{k}$, with two unit vectors ${\bf e}_j \in \mathbb{P}^{(i)}_{1}$ and ${\bf e}_k \in \mathbb{P}^{(i)}_{2}$. When considering $\mathbb{P}^{(i)}_1 \subseteq \mathbb{R}^{m}$ and $\mathbb{P}^{(i)}_2 \subseteq \mathbb{R}^{n}$, the vector space $\mathbb{P}^{(i)}$ composed by $\mathbb{P}^{(i)}_{1}$ and $\mathbb{P}^{(i)}_{2}$ should be contained in $\mathbb{R}^{m\times n}$. This composition is mathematically represented as $\mathbb{P}^{(i)}_1 \otimes \mathbb{P}^{(i)}_2 \subseteq \mathbb{R}^{m \times n}$, where $\otimes$ refers to direct product.

The above statement involves two linear operations: direct sum ($\oplus$) and direct product ($\otimes$), which result in vector spaces contained in $\mathbb{R}^{m+n}$ and $\mathbb{R}^{m \times n}$, respectively. However, direct sum and direct product represent two different perspectives of the same process: the composition of a vector from two subspaces. As summarized in~\cref{tab:op}, direct sum is a kind of coordinate extension in the Cartesian coordinate system, with the represented range by the newly formed basis set in Euclidean space as a scalar multiplied as described by direct product. Since we have diminished the dimensionality in our parameterization in~\cref{def:vec}, we use the vectors in the Cartesian coordinate system to address the composition afterwards. In this paper, to avoid any confusion, we use $\mathbb{P}_1 \times \mathbb{P}_2 \rightarrow \mathbb{P}$ to represent a vector space composition process from $\mathbb{P}_1$ and $\mathbb{P}_2$ to $\mathbb{P}$. During this process, the vector coordinate change is $\text{dim}(\mathbb{P}_1 \times \mathbb{P}_2) = \text{dim}(\mathbb{P}_1) + \text{dim}(\mathbb{P}_2)$ and the represented range change by the vector is $\text{dim}(\mathbb{P}_1 \times \mathbb{P}_2)= \text{dim}(\mathbb{P}_1)\text{dim}(\mathbb{P}_2)$., where $\text{dim}(\cdot)$ is the vector coordinate length (vector dimension) of a vector space.

\begin{table}[]
\centering
\begin{tabular}{c|c|c|c|c}

\toprule
{\bf Operation} & \begin{tabular}[c]{c}{\bf Dimension}\\ {\bf change}\end{tabular}& {\bf Geometric meaning} & \begin{tabular}[c]{c}{\bf Order}\\ {\bf change}\end{tabular} & \begin{tabular}[c]{c}{\bf Occurrence in this paper} \end{tabular} \\ \midrule
Direct Sum & $\mathbb{R}^{m} \times \mathbb{R}^{n} \rightarrow \mathbb{R}^{m+n}$ & \begin{tabular}[l]{l}Coordinate/dimension \\ extension of vectors\end{tabular} & $0$ & {\begin{tabular}[l]{l}\cref{def:mat-comp,def:ten-comp} \\ \cref{fig:add,eq:v-idx}\end{tabular}} \\ \hline

\begin{tabular}[c]{c}{Direct Product} \\ {= Tensor Product} \end{tabular} &  $\mathbb{R}^{m} \times \mathbb{R}^{n}\rightarrow \mathbb{R}^{m\times n} $  & \begin{tabular}[c]{c}{Order extension of vectors,}\\ {or representative range} \\{expansion of vector space}\end{tabular}    &  $+1$       &  \begin{tabular}[c]{c}\cref{def:mat-comp,def:ten-comp} \\ \cref{fig:matmul,fig:matmul-proj}\\ \cref{eq:kron,th:tsvd}\end{tabular}     \\ \hline
Flatten &   $\operatorname{vec}(\mathbb{R}^{m\times n}) \rightarrow \mathbb{R}^{mn}  $    & \begin{tabular}[l]{l}Hypercube volume\\ / calculus of dimensions\end{tabular} &   $-1$       &  \begin{tabular}[c]{c} \cref{th:svd,th:tsvd} \\ \cref{eq:kron} \end{tabular}
\\ \bottomrule
\end{tabular}
\caption{Operations on subspaces (i.e. composition or decomposition) in the Euclidean space. In the context of the Cartesian coordinate system, direct sum and direct product result in the same vector space, though direct sum emphasises vector coordinates and direct product emphasises the vector order and dimension. In the following text, we use $\times$ between vector spaces (and the vectors in these vector spaces) to represent a composed vector space (a vector formed by the other two vectors), and $\otimes$ only refers to the Kronecker product. In~\cref{sec:decompose}, when we use a vector to represent tensors, tensor product (and even tensor contraction) is prevalent with the direct product.}\label{tab:op}
\end{table}

\begin{definition}[Matrix Composition in Euclidean Space]\label{def:mat-comp} A matrix ${\bf C} \in \mathbb{R}^{m\times n}$ can be represented with two orthogonal vectors ${\bf a}$, ${\bf b}$ in a vector space $\mathbb{P} = \mathbb{R}^{m+n}$, with ${\bf a} = (a_1, a_2, \ldots, a_{m}, \overbrace{0,\ldots, 0)}^{n}$, ${\bf b} = (\overbrace{0,\ldots, 0}^{m}, b_1, b_2, \ldots, b_{n})$. Therefore, the $(i,j)$th element of ${\bf C}$ is $c_{i,j} = a_{i} b_{n+j}$, where $i, j < m+n$. By reducing the numbers of the elements as $1 \leq i \leq n$ and $1+n \leq j \leq m+n$, we obtain the final matrix representation as:
\begin{equation}\label{eq:map}
    {\bf C} = {\bf a} \times {\bf b}=
\left[\begin{array}{cccc}
        a_1 b_1 & a_1 b_2 & \ldots & a_1 b_n \\ 
        a_2 b_1 & a_2 b_2 & \ldots & a_2 b_n  \\
        \vdots & \vdots & \ddots & \vdots \\
        a_m b_1 & a_m b_2 & \ldots & a_m b_n \\
\end{array}\right]
\in \mathbb{R}^{m\times n}
\end{equation}
    
\end{definition}

We use two $(m+n)$-dimensional vectors ${\bf a}$ and ${\bf b}$ to represent a matrix ${\bf C} \in \mathbb{R}^{m\times n}$ in~\cref{def:mat-comp}. An attractive property for this kind of representation is that ${\bf a}$ and ${\bf b}$ are in the same subspace, $\mathbb{R}^{m+n}$, which has fewer dimensions than $\mathbb{R}^{m\times n}$. In doing this, we can describe their interactions in the whole parameter space $\mathbb{P}$ and interpret common language model modules in~\cref{sec:lm-module},  which we cannot achieve when putting two vectors in disjoint subspaces $\mathbb{R}^m$ and $\mathbb{R}^n$.

\begin{figure}[h!]
    \centering
    \begin{subfigure}[b]{\textwidth}
        \centering
        \begin{tikzpicture}

    \node[color=RoyalBlue] at (-0.8,0,-0.25) {$\mathbb{P}_1 \subset \mathbb{R}^{m}$};
    \node[color=gray] at (3.5,0.8,0) {${\bf W}^{\top}$};
    \node[color=Maroon] at (5.25,0,-0.25) {$\mathbb{P}_2\subset \mathbb{R}^{n}$};
    
    \draw[ultra thick,-stealth] (0,0,0) -- (1,0.75,1) node[anchor=south]{$\bf{x}$};
    \draw[line width=0.2mm, -stealth,RoyalBlue] (0,0,0) -- (1,0,0) node[anchor=north east]{};
    \draw[line width=0.2mm,-stealth,RoyalBlue] (0,0,0) -- (0,1,0) node[anchor=north west]{};
    \draw[line width=0.2mm,-stealth,RoyalBlue] (0,0,0) -- (0,0,1) node[anchor=east]{};

    \draw[line width=0.2mm, -stealth,RoyalBlue, dashed] (3,0,0) -- (4,0,0) node[anchor=north east]{};
    \draw[line width=0.2mm,-stealth,RoyalBlue, dashed] (3,0,0) -- (3,1,0) node[anchor=north west]{};
    \draw[line width=0.2mm,-stealth,RoyalBlue, dashed] (3,0,0) -- (3,0,1) node[anchor=east]{};
    \draw[line width=0.2mm, -stealth,Maroon, dashed] (3,0,0) -- (3.5,-0.25,-0.25) node[anchor=north east]{};
    \draw[line width=0.2mm,-stealth,Maroon, dashed] (3,0,0) -- (3.5,0.5,0) node[anchor=north west]{};
    \draw[line width=0.2mm,-stealth,Maroon, dashed] (3,0,0) -- (2.5,0,1) node[anchor=east]{};

    \draw[ultra thick, -stealth, gray] (3.2,1,0) .. controls  (4.25,2,1) and (4.25,1.5,1) .. (3.7,0.5,0);

    \draw[line width=0.2mm, -stealth,Maroon] (6,0,0) -- (6.5,-0.25,-0.25) node[anchor=north east]{};
    \draw[line width=0.2mm,-stealth,Maroon] (6,0,0) -- (6.5,0.5,0) node[anchor=north west]{};
    \draw[line width=0.2mm,-stealth,Maroon] (6,0,0) -- (5.5,0,1) node[anchor=east]{};

    \draw[ultra thick,-stealth] (6,0,0) -- (6.5,-0.25,0.5) node[anchor=west]{$\bf{y}$};
    
\end{tikzpicture}
        \caption{Linear mapping view of matrix-vector multiplication ${\bf x}{\bf W}^{\top}= {\bf y}$, ${\bf W} \in \mathbb{R}^{m \times n}$ and ${\bf W}^{\top}: \mathbb{R}^{m} \rightarrow \mathbb{R}^{n}$}
        \label{fig:matmul}
    \end{subfigure}
    \hfill
    \begin{subfigure}[b]{\textwidth}
        \centering
        \begin{tikzpicture}

    \node[color=RoyalBlue] at (-0.8,0,-0.25) {$\mathbb{P}_1 \subset \mathbb{R}^{m}$};
    \node[color=Maroon] at (5.25,0,-0.25) {$\mathbb{P}_2\subset \mathbb{R}^{n}$};
    
    \draw[ultra thick,-stealth] (0,0,0) -- (1,0.75,1) node[anchor=south]{$\bf{x}$};
    \draw[ultra thick,-stealth] (0,0,0) -- (-0.5,1,0) node[anchor=south]{$\bf{a}$};
    
    \coordinate (x) at (0.5,0.75/2,0.5);
    \coordinate (a) at (-0.25,0.5,0);
    \coordinate (O) at (0,0,0);
    \pic [draw, "$\theta$", angle radius=0.2cm, angle eccentricity=2] {angle = x--O--a};
    
    \draw[line width=0.2mm, -stealth,RoyalBlue] (0,0,0) -- (1,0,0) node[anchor=north east]{};
    \draw[line width=0.2mm,-stealth,RoyalBlue] (0,0,0) -- (0,1,0) node[anchor=north west]{};
    \draw[line width=0.2mm,-stealth,RoyalBlue] (0,0,0) -- (0,0,1) node[anchor=east]{};

    \draw[line width=0.2mm, -stealth,RoyalBlue] (3,0,0) -- (4,0,0) node[anchor=north east]{};
    \draw[line width=0.2mm,-stealth,RoyalBlue] (3,0,0) -- (3,1,0) node[anchor=north west]{};
    \draw[line width=0.2mm,-stealth,RoyalBlue] (3,0,0) -- (3,0,1) node[anchor=east]{};

    \draw[ultra thick,-stealth, opacity=.5] (3,0,0) -- (4,0.75,1) node[anchor=south]{$\bf{x}$};
    \draw[ultra thick,-stealth, opacity=.5] (3,0,0) -- (2.5,1,0) node[anchor=south]{$\bf{a}$};
    \draw[thick, opacity=.5] (2.5,1,0) -- (4,0.75,1);
    \node at (3.8,1,0) {$\textcolor{RoyalBlue}{a} = \langle {\bf x}, {\bf a}\rangle$};
    \coordinate (x2) at (3.5,0.75/2,0.5);
    \coordinate (a2) at (-0.25+3,0.5,0);
    \coordinate (O2) at (3,0,0);
    \pic [draw, "$\theta$", angle radius=0.2cm, angle eccentricity=2] {angle = x2--O2--a2};



    \draw[line width=0.2mm, -stealth,Maroon] (6,0,0) -- (6.5,-0.25,-0.25) node[anchor=north east]{};
    \draw[line width=0.2mm,-stealth,Maroon] (6,0,0) -- (6.5,0.5,0) node[anchor=north west]{};
    \draw[line width=0.2mm,-stealth,Maroon] (6,0,0) -- (5.5,0,1) node[anchor=east]{};

    \draw[ultra thick,-stealth] (6,0,0) -- (6.5,-0.25,0.5) node[anchor=west]{${\bf y}=\textcolor{RoyalBlue}{a}{\bf b}$};

    \draw[ultra thick,-stealth, black!40] (6,0,0) -- (6.25,-0.25/2,0.25) node[anchor=east]{$\bf{b}$};
    
\end{tikzpicture}
        \caption{Projection view of matrix-vector multiplication ${\bf x}{\bf W}^{\top}= {\bf y}$, ${\bf W}^{\top} = {\bf a}\times {\bf b}$, where ${\bf a}, {\bf x} \in \mathbb{R}^{m}$ and ${\bf b} \in \mathbb{R}^{n}$}
        \label{fig:matmul-proj}
    \end{subfigure}
    \hfill
        \begin{subfigure}[b]{\textwidth}
        \centering
        \begin{tikzpicture}

    \node[color=RoyalBlue] at (-0.8,0,-0.25) {$\mathbb{P}_1 \subset \mathbb{R}^{n}$};
    \node[color=Maroon] at (1.9,0,-0.25) {$\mathbb{P}_2\subset \mathbb{R}^{m}$};
    \node[color=Orchid] at (4.6,0,0) {$\mathbb{P}_1 \oplus \mathbb{P}_2$};
    \node[color=gray] at (3.7,0.65,0) {$\oplus$};
    
    \draw[ultra thick,-stealth] (0,0,0) -- (1,0.75,1) node[anchor=south]{$\bf{u}$};
    \draw[line width=0.2mm, -stealth,RoyalBlue] (0,0,0) -- (1,0,0) node[anchor=north east]{};
    \draw[line width=0.2mm,-stealth,RoyalBlue] (0,0,0) -- (0,1,0) node[anchor=north west]{};
    \draw[line width=0.2mm,-stealth,RoyalBlue] (0,0,0) -- (0,0,1) node[anchor=east]{};

    \draw[line width=0.2mm, -stealth,Maroon] (2.5,0,0) -- (3,-0.25,-0.25) node[anchor=north east]{};
    \draw[line width=0.2mm,-stealth,Maroon] (2.5,0,0) -- (3,0.5,0) node[anchor=north west]{};
    \draw[line width=0.2mm,-stealth,Maroon] (2.5,0,0) -- (2,0,1) node[anchor=east]{};

    \draw[ultra thick,-stealth] (2.5,0,0) -- (3,-0.25,0.5) node[anchor=west]{$\bf{v}$};

    \draw[ultra thick, -stealth, gray] (3.2,0.4,0) -- (4.2,0.4,0);

    \draw[line width=0.2mm, -stealth,Orchid] (5.5,0,0) -- (6.5,0,0) node[anchor=north east]{};
    \draw[line width=0.2mm,-stealth,Orchid] (5.5,0,0) -- (5.5,1,0) node[anchor=north west]{};
    \draw[line width=0.2mm,-stealth,Orchid] (5.5,0,0) -- (5.5,0,1) node[anchor=east]{};
    \draw[line width=0.2mm, -stealth,Orchid] (5.5,0,0) -- (6,-0.25,-0.25) node[anchor=north east]{};
    \draw[line width=0.2mm,-stealth,Orchid] (5.5,0,0) -- (6,0.5,0) node[anchor=north west]{};
    \draw[line width=0.2mm,-stealth,Orchid] (5.5,0,0) -- (5,0,1) node[anchor=east]{};

    \draw[thick] (5.5,0,0) -- (6.5,0.75,1) node[anchor=south east]{$\bf{u}$};
    \draw[thick] (6.5,0.75,1) -- (7,0.5,1.5) node[anchor=south west]{$\bf{v}$};
    \draw[ultra thick,-stealth] (5.5,0,0) -- (7,0.5,1.5) node[anchor=north east]{${\bf u} + {\bf v}$};
    
\end{tikzpicture}
        \caption{${\bf u} + {\bf v}$ or concantenation $\texttt{concat}({\bf u}, {\bf v})$, $\mathbb{P}_1 \bot \mathbb{P}_2$ and $\mathbb{P}_1 \times \mathbb{P}_2 \subseteq \mathbb{R}^{n+m}$ (regarding coordinates).} 
        \label{fig:add}
    \end{subfigure}
\hfill
        \begin{subfigure}[b]{\textwidth}
        \centering
        \begin{tikzpicture}

    \node[color=RoyalBlue] at (-0.8,0,-0.25) {$\mathbb{Q} \subset \mathbb{R}^{d_k}$};
    \node[color=Maroon] at (2.4,0.2,-0.25) {$\mathbb{K}\subset \mathbb{R}^{d_k}$};
    \node[color=Maroon] at (5.5,0.2,-0.25) {$\mathbb{K}\subset \mathbb{R}^{d_k}$};
    \node[color=OliveGreen] at (8.2,0.2,-0.25) {$\mathbb{V} \subset \mathbb{R}^{d_v}$};
    \node[color=gray] at (1.5,0.65,0) {${\bf K}^{\top}$};
    \node[color=gray] at (4.3,0.65,0) {$\texttt{softmax}$};
    \node[color=gray] at (4.3,0.1,0) {$\frac{1}{\sqrt{d_k}}$};
    \node[color=gray] at (7.3,0.65,0) {${\bf V}$};
    
    \draw[ultra thick,-stealth] (0,0,0) -- (1,0.75,1) node[anchor=south]{$\bf{q}$};
    \draw[line width=0.2mm, -stealth,RoyalBlue] (0,0,0) -- (1,0,0) node[anchor=north east]{};
    \draw[line width=0.2mm,-stealth,RoyalBlue] (0,0,0) -- (0,1,0) node[anchor=north west]{};
    \draw[line width=0.2mm,-stealth,RoyalBlue] (0,0,0) -- (0,0,1) node[anchor=east]{};

    \draw[ultra thick,-stealth] (3,0.2,0) -- (3.5,-0.05,0.5) node[anchor=west]{$\bf{k}$};

    \draw[ultra thick, -stealth, gray] (1,0.4,0) -- (1.9,0.4,0);

    \draw[line width=0.2mm, -stealth,Maroon] (3,0.2,0) -- (3.5,-0.05,-0.25) node[anchor=north east]{};
    \draw[line width=0.2mm,-stealth,Maroon] (3,0.2,0) -- (3.5,0.7,0) node[anchor=north west]{};
    \draw[line width=0.2mm,-stealth,Maroon] (3,0.2,0) -- (2.5,0.2,1) node[anchor=east]{};

    \draw[ultra thick,-stealth] (6.2,0.2,0) -- (6.5,0.1,0.7) node[anchor=west]{$\tilde{\bf k}$};

    \draw[line width=0.2mm, -stealth,Maroon] (6.2,0.2,0) -- (6.7,-0.05,-0.25) node[anchor=north east]{};
    \draw[line width=0.2mm,-stealth,Maroon] (6.2,0.2,0) -- (6.7,0.7,0) node[anchor=north west]{};
    \draw[line width=0.2mm,-stealth,Maroon] (6.2,0.2,0) -- (5.7,0.2,1) node[anchor=east]{};
    
    \draw[ultra thick, -stealth, gray] (3.7,0.4,0) -- (5,0.4,0);

    \draw[line width=0.2mm, -stealth,Maroon] (6.2,0.2,0) -- (6.7,-0.05,-0.25) node[anchor=north east]{};
    \draw[line width=0.2mm,-stealth,Maroon] (6.2,0.2,0) -- (6.7,0.7,0) node[anchor=north west]{};
    \draw[line width=0.2mm,-stealth,Maroon] (6.2,0.2,0) -- (5.7,0.2,1) node[anchor=east]{};
    
    \draw[ultra thick, -stealth, gray] (6.9,0.4,0) -- (7.7,0.4,0);

    \draw[line width=0.2mm, -stealth,OliveGreen] (9.2,0.2,0) -- (10,0.5,1) node[anchor=north east]{};
    \draw[line width=0.2mm,-stealth,OliveGreen] (9.2,0.2,0) -- (9,0.7,-1) node[anchor=north west]{};
    \draw[line width=0.2mm,-stealth,OliveGreen] (9.2,0.2,0) -- (8.7,0.2,1) node[anchor=east]{};

        \draw[ultra thick,-stealth] (9.2,0.2,0) -- (8.8,0.7,0.1) node[anchor=east]{${\bf v}$};

\end{tikzpicture}
        \caption{Dot-product attention ${\bf a} = \texttt{softmax}(\frac{{\bf q}{\bf K}^{\top}}{\sqrt{d_k}}){\bf V}$~\cite{vaswani2017attention}}
        \label{fig:attn}
    \end{subfigure}
    
    \caption{Typical geometric transformations in transformer models.}
    \label{fig:geo-lm}
\end{figure}
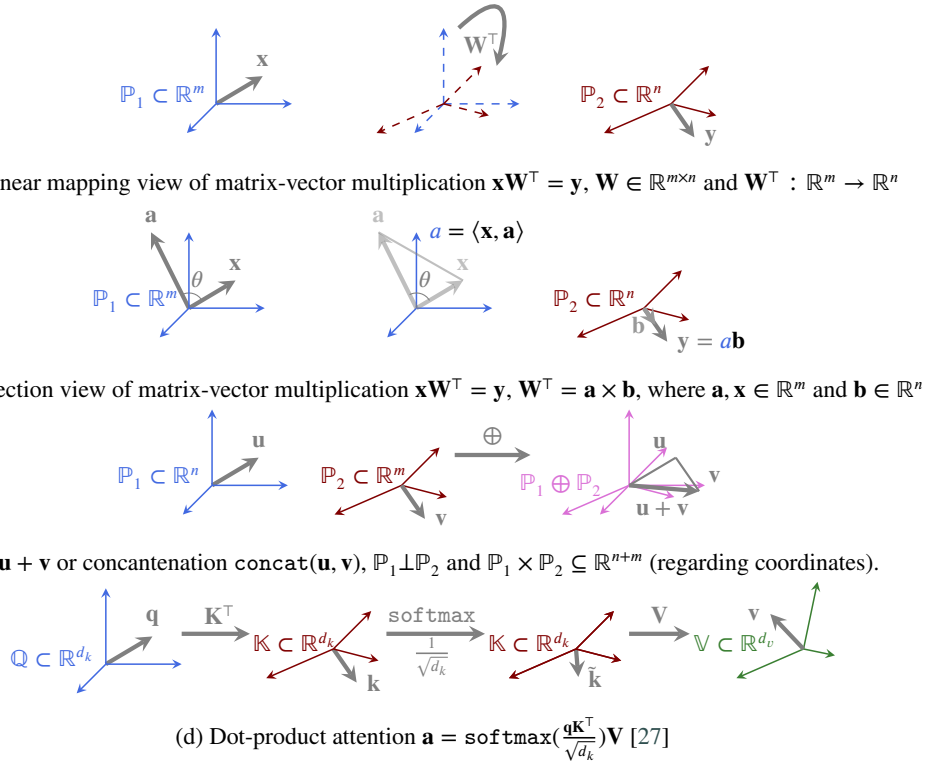

The geometric intuition of matrix composition is visualized in~\cref{fig:matmul}. Suppose ${\mathbb{P}}$ has a basis set $({\bf v}_1, {\bf v}_2, \ldots, {\bf v}_m, \\{\bf w}_1, {\bf w}_2, \ldots, {\bf w}_n) $, so that ${\bf a} = \sum^{m}_{i=1} a_{i}{\bf v}_i$ and ${\bf b} = \sum^{n}_{i=1} b_{i}{\bf w}_i$. The subspace ${\mathbb{P}_1}$ is defined with the base $({\bf v}_1, {\bf v}_2, \ldots, {\bf v}_m)$, and subspace ${\mathbb{P}_2}$ is defined with the base $({\bf w}_1, {\bf w}_2, \ldots, {\bf w}_n)$, thus ${\bf a} \in \mathbb{P}_1$ and ${\bf b} \in \mathbb{P}_2$. Denote a linear transformation ${\bf C}: \mathbb{P}_1 \rightarrow \mathbb{P}_2$, such that ${\bf C}{\bf v}_{i} = \sum^{m}_{j=1} c_{ij}{\bf w}_{j}$. Here, matrix ${\bf C}$ acts like a dictionary mapping any vector from $\mathbb{P}_1$ to a certain vector in $\mathbb{P}_2$. During this linear transformation, vector ${\bf a}$ is the original \ul{operand}, and vector ${\bf b}$ defines the actual transformation ${\bf C}$, or in other words, \ul{operation}.

\begin{remark} Matrix composition defined in~\cref{def:mat-comp} can be seen as a process of transforming a vector ${\bf a} \in \mathbb{P}_1$ into another vector ${\bf c} \in \mathbb{P}_2$, where the transformation is defined by ${\bf b}$. Inversely, there exists a linear transformation ${\bf C}^{\top}: \mathbb{P}_2 \rightarrow \mathbb{P}_1$, with a mapping matrix as $ {\bf C}^{\top}$.
\end{remark}

The procedure of defining the matrix with two subspaces in~\cref{def:mat-comp} can be easily extended to tensors.

\begin{definition}[Tensor Composition in Euclidean Space]\label{def:ten-comp} An $n$-order tensor $\mathcal{A} \in \mathbb{R}^{d_1 \times d_2 \times \cdots \times d_n}$ can be represented as the $n$ orthogonal vectors ${\bf v} = (v_1, v_2, \ldots, v_{d_1}, \overbrace{0, \ldots, 0}^{\sum^{n}_{i=2}d_i})$, ${\bf w} = (\overbrace{0,\ldots, 0}^{d_1}, w_1, w_2, \ldots, w_{d_2}, \overbrace{0, \ldots, 0 }^{\sum^{n}_{i=3}d_i})$, $\ldots$ , ${\bf u} = (\overbrace{0,\ldots, 0}^{\sum^{n-1}_{i=1}d_i}, u_1, u_2, \ldots, u_{d_n})$, where its $(x, y, \ldots, z)$th entry is 
$ \mathcal{A}[x, y, \ldots, z] = \overbrace{v_{x}w_{y+d_1}\cdots u_{z+\sum^{n-1}_{i=1} d_i} }^{n \; \text{elements}}$. Here, ${\bf v} \in \mathbb{P}_1 \subset \mathbb{R}^{d_1}$, ${\bf w} \in \mathbb{P}_2 \subset \mathbb{R}^{d_2}$, $\cdots$, ${\bf u} \in \mathbb{P}_n\subset \mathbb{R}^{d_n}$. 
\end{definition}
In~\cref{def:ten-comp}, there are $n$ subspaces, and these subspaces construct an $n$-order tensor. The parameter space $\mathbb{P} \in \mathbb{R}^{d_1+d_2+\cdots+d_n}$ can therefore represent tensors in $\mathbb{R}^{d_1\times d_2 \times\cdots\times d_n}$.

\subsection{Interpretion of Transformer Modules with Subspaces}\label{sec:lm-module}

In this section, we give the geometric interpretation of common transformer modules, given in~\cref{def:mat-comp,def:ten-comp}, and graphically illustrated in~\cref{fig:lm-mc}. Transformer structure typically maps the outputs of embedding layers to a subspace of similar dimension, subsequently maps the output features from the attention layer to a higher dimensional latent subspace (graphically represented by bigger cubes), and finally to lower dimensional subspace (graphically represented by smaller cubes). For a multi-head transformer structure, this sequential mapping is executed multiple times in parallel, which is graphically represented as replicated groups of ``Attn''(attention layer) and ``FNN''(feed-forward layer) modules.

\subsubsection{Linear Layer (the only parametric part of feed-forward layer)}
A linear layer, also called a fully connected layer, dense layer, or affine layer, is the only component having weights in the feed-forward layer. Its output, ${\bf y} \in \mathbb{R}^n$, and input, ${\bf x} \in \mathbb{R}^m$, satisfy
\begin{equation}
    {\bf y}={\bf x}{\bf W}^{\top}+ {\bf c}, 
\end{equation}
where ${\bf W}\in \mathbb{R}^{n\times m}$ is the weight matrix, and ${\bf c}\in \mathbb{R}^n$ is the bias. For simplicity, we only discuss the weight matrix rather than weight tensors here. 

The forward pass of the linear layer represents basically matrix multiplication or vector-matrix multiplication; these two are treated as equivalent in this paper.
A common geometric interpretation of such process is shown in~\cref{fig:matmul}, where the vector-matrix multiplication between vector ${\bf x}$ and matrix ${\bf W}^{\top}$ can be seen as the vector ${\bf x}$ being projected from the original subspace $\mathbb{P}_1$ to another subspace $\mathbb{P}_2$, where $\mathbb{P}_1, \mathbb{P}_2 \subset \mathbb{P}$.

However, based on the matrix composition defined in~\cref{def:mat-comp}, such transformation can be interpreted as two  operations shown in~\cref{fig:matmul-proj}: $\large{\textcircled{\small 1}}$ ${\bf x}, {\bf a} \in \mathbb{P}_1$,  $\langle {\bf x}, {\bf a}\rangle \rightarrow a \in \mathbb{R}$, $\large{\textcircled{\small 2}}$ $a{\bf b}, {\bf b} \in \mathbb{P}_2$, where $\langle \cdot,\cdot \rangle$  is inner product (projection product in sense of geometry), and ${\bf a}$ and ${\bf b}$ are the vectors composing weight matrix ${\bf W}^{\top}$. 

\subsubsection{Concantenation}\label{sec:concat}
Concantenation in the context of generative langauge models usually occurs when merging the outputs of the transformer heads (i.e. numeric vectors) by simply append the outputs one-by-one. Since there are no arithmetic interactions among the elements in these vectors, this process can be seen as the merge of two orthogonal subspaces with two disjoint basis sets. This merging can be easily implemented via a construct a larger subspace $\mathbb{P}_3 = \mathbb{P}_1 \times \mathbb{P}_2$, as shown in~\cref{fig:add}.     

\subsubsection{Attention Layer}

Attention layers are the core functional modules in the transformer-based language models. Take the original transformer architecture~\cite{vaswani2017attention} as an example, the main mathematical expressions of the attention and head (a module of the neural network) is as following
\begin{equation}\label{eq:attn_vec}
    \texttt{attn}({\bf q},{\bf K},{\bf V}) = \texttt{softmax}\frac{{\bf q}{\bf K}^{\top}}{\sqrt{d_k}}{\bf V},
\end{equation}
where ${\bf q}$ a the query vector, ${\bf K}$ and ${\bf V}$ are respectively the key and value matrices, and $d_k$ is the dimension of ${\bf q}$. This transformation process can be geometrically interpreted as in~\cref{fig:attn}. The query ${\bf q}$ is first projected into the subspace $\mathbb{K}$, which represents keys. Then, the resulting vector ${\bf k}$ is normalized via ${\texttt{softmax}}$ and the coefficient $\frac{1}{\sqrt{d_k}}$ in the same subspace $\mathbb{K}$, to become another vector  $\tilde{\bf k}$. Finally, the normalized vector, $\tilde{\bf k}$, is projected to the value subspace, $\mathbb{V}$. 

Therefore, the attention layer is a series of linear transformations within three subspaces: query subspace $\mathbb{Q}$, key subspace $\mathbb{K}$ and value subspace $\mathbb{V}$, and its utimate goal is to mapping a query vector to a value, and this value is named ``attention''. Or in other word, the attention layer is a series of two projections, project a single query ${\bf q}$ to a value.

\section{Matrix and Tensor Factorization from a Subspace Composition View}\label{sec:decompose}
In this section, we provide the specific details of the relevant methodologies - matrix and tensor factorization. All the descriptions will be based on the matrix composition in~\cref{def:mat-comp} or tensor composition in~\cref{def:ten-comp}. However, before discussing factorizations in detail, we should clarify typical products from subspace, so as to avoid confusion. Some of these products are summarized in~\cref{tab:op}.

\subsection{Common Operations on Subspaces}
We have discussed the operations on subspaces in~\cref{tab:op}. In this section, we provide more clarification on this in light of matrix factorization (\cref{sec:matd}) and tensor factorization (\cref{sec:td}).

\subsubsection{Cartesian Product and Tensor Product on $1$-order Subspaces }\label{sec:1-order-prod}
Taking a vector space $\mathbb{V} \subseteq \mathbb{R}^{d_1 \times d_2 \times \cdots \times d_n}$ as an example, it is an $n$-order vector space constructed by $n$ $1$-order subspaces $\mathbb{V}_1 \subseteq \mathbb{R}^{d_1}$, $\mathbb{V}_2 \subseteq \mathbb{R}^{d_2}$, $\ldots$, $\mathbb{V}_n \subseteq \mathbb{R}^{d_n}$.
The basis set of $\mathbb{V}$ is $\mathcal{V}$, 
$\mathcal{V}=\mathcal{V}_1 \times \mathcal{V}_2 \times \cdots \times \mathcal{V}_n$
, where $\mathcal{V}_i$ is the basis of $\mathbb{V}_i$, $\vert \mathcal{V}_i \vert = \text{dim}({\mathbb{V}_i}) =d_i$. Similar to what we discussed in~\cref{sec:subspaces}, we use the notation $\mathbb{V}_1 \times \mathbb{V}_2 \times \cdots \times \mathbb{V}_n = \mathbb{V}$ to represent the process of constructing a new vector space $\mathbb{V}$ upon $\mathbb{V}_1, \mathbb{V}_2, \ldots, \mathbb{V}_n$.
This process is officially called {\it Cartesian product}, which means any vector ${\bf v}\in \mathbb{V}$ can be represented as a linear combination of the orthonormal vectors in $\mathcal{V}$. Specifically for~\cref{eq:v-idx}, ${\bf v} = (v^{(1)}_1, v^{(1)}_2, \ldots, v^{(1)}_{d_1}, \textcolor{cyan}{\big|} v^{(2)}_1,v^{(2)}_2,\ldots, \textcolor{cyan}{\big|} v^{(n)}_1,v^{(n)}_2,\ldots, v^{(n)}_{d_n})$, where the length of the vector representation is $\vert \mathcal{V} \vert = \text{dim}(\mathbb{V}) =\sum^{n}_{i=1} d_i$, so that the representation range of the vector ${\bf v}$ grows to $\prod^{n}_{i=1} d_i$

{\it Tensor product} is a widely used term for such dimension increase caused by ``order concatenation'', whose operands are order-$1$ vectors, or in other words, vectors in order-$1$ subspaces. Obviously, direct product of subspaces in~\cref{tab:op} is a generalized version of the tensor product, and in the following texts we refer to both cases as ``product'' with notation $\times$. 

\subsubsection{Product and Contraction on $n$-order Subspaces}\label{sec:n-order-prod}
Similar to $\mathbb{V}$ in~\cref{sec:1-order-prod}, suppose there is another order-$m$ vector space $\mathbb{W} \subseteq \mathbb{R}^{h_1 \times h_2 \times \cdots \times h_m}$ consists of subspaces $\mathbb{W}_1 \subseteq \mathbb{R}^{h_1}$, $\mathbb{W}_2 \subseteq \mathbb{R}^{h_2}$, $\ldots$, $\mathbb{W}_m \subseteq \mathbb{R}^{h_m}$, with a basis $\mathcal{W}=\mathcal{W}_1 \times \mathcal{W}_2 \times \cdots \times \mathcal{W}_m$. $\mathcal{W} \cap \mathcal{V} =\varnothing$. 
Then, the tensor product of $\mathbb{V}$ and $\mathbb{W}$ is the same as order-$1$ tensor product in~\cref{sec:1-order-prod}, such that $\mathbb{V} \times \mathbb{W}= \mathbb{V}_1 \times \cdots \times \mathbb{V}_n \times \mathbb{W}_1 \times \cdots \times \mathbb{W}_m$, with a basis set $(\mathcal{V} \times \mathcal{W}) = \mathcal{V}_1 \times \cdots \times \mathcal{V}_n \times \mathcal{W}_1 \times \cdots \times \mathcal{W}_m $.

{\it Contraction} can be seen as a general version of tensor product. For the vector spaces $\mathbb{V}$ and $\mathbb{W}$, suppose $\vert \mathcal{V}_i \vert = \vert \mathcal{W}_j \vert$. Then, the mode-$(i,j)$ contraction of $\mathbb{V}$ and $\mathbb{W}$ yields $\mathbb{V}_1 \times \cdots \times \mathbb{V}_{i-1} \times \mathbb{V}_{i+1} \times \cdots \times \mathbb{V}_n \times \mathbb{W}_1 \times \cdots \times \mathbb{W}_{j-1} \times \mathbb{W}_{j+1} \times \cdots \times \mathbb{W}_m$. In short, contraction ``absorbs'' the common subspaces shared by $\mathbb{V}$ and $\mathbb{W}$.
Then we have the cardinality of the basis set after tensor product/contraction of $\mathbb{V}$ and $\mathbb{W}$ is

\begin{align}\label{eq:prod-basis}
\vert \mathcal{V} \times \mathcal{W} \vert = \vert \mathcal{V} \cup \mathcal{W} \vert -  \vert \mathcal{V} \cap \mathcal{W} \vert = \left\{ \begin{array}{lll} \vert \mathcal{V} \vert + \vert \mathcal{W} \vert
, & {\mathcal{V} \cap \mathcal{W} = \varnothing}, & \text{tensor product},  \\
\vert \mathcal{V} \vert + \vert \mathcal{W} \vert - 2 \vert \mathcal{V} \cap \mathcal{W} \vert , & {\mathcal{V} \cap \mathcal{W} \neq \varnothing, } & \text{tensor contraction}.
\end{array} \right.
\end{align}

\subsection{Matrix Factorization}\label{sec:matd} 
Among different matrix factorization algorithms in this section, we only consider Singular Value Decomposition (SVD) and Kronecker Product. 

\subsubsection{Singular Value Decomposition(SVD) and Matrix (Cayley) Multiplication}\label{th:svd}
We refer to the following representation from~\citep[page 275, 277]{axler2024linear} for a clear interpretation of SVD via subspaces.

\begin{theorem}[Singular Value Decomposition~\citep{axler2024linear}]\label{th:svd} Suppose an invertible linear transformation $T: {\mathbb{P}_1} \rightarrow {\mathbb{P}_2}$, and ${\bf v}=v_1 {\bf v}_1 + v_2 {\bf v}_2 + \cdots + v_m {\bf v}_m$, ${\bf v}  \in \mathbb{P}_1$, ${\bf w}=w_1 {\bf w}_1 + w_2 {\bf w}_2 + \cdots + w_m {\bf w}_m$, ${\bf w}  \in \mathbb{P}_2$. Then, we have 
\begin{equation}\label{eq:svd}
    {\bf T}{\bf v} = {\bf v} {\bf T}^{\top} = \sum^{m}_{i=1} s_i v_i {\bf w}_i, 
\end{equation}
for every ${\bf v}\in \mathbb{P}_1$ and ${\bf w}\in \mathbb{P}_2$.
\end{theorem}

In~\cref{th:svd}, ${\bf v}_i$ and ${\bf w}_i$ are unit vectors in $\mathbb{P}_1$ and $\mathbb{P}_2$ respectively, and $s_1, s_2, \ldots, s_m$ are the singular values of ${\bf T}$. In this way, the linear transformation of ${\bf v}$ from $\mathbb{P}_1$ to $\mathbb{P}_2$ can be represented with $m$ orthogonal vectors in $\mathbb{P}_2$. The geometric illustration is exactly the same as that in~\cref{fig:matmul-proj}, where $\langle {\bf x}, {\bf a}\rangle \rightarrow a $, ${\bf x}W^{\top}=a{\bf b}$ changes to $\langle {\bf v}, {\bf v}_i\rangle \rightarrow v_i$, ${\bf v}_i {\bf T}^{\top}= s_i{\bf w}_i \Rightarrow {\bf v} {\bf T}^{\top}= \sum^{m}_{i=1} v_i (s_i{\bf w}_i)$.

We can observe in~\cref{eq:svd} for $i$-axis in $\mathbb{P}_2$, singular values $s_i$ are the part of the coefficients of the unit vectors ${\bf w}_i$. If ${\bf w}_i$ is small enough, we can ignore this orthogonal vector $s_i v_i {\bf w}_i$. If we ignore $(m-k)$ orthogonal vectors in~\cref{eq:svd}, we can further reduce the representation of result vector in $\mathbb{P}_2$ from $m$ dimension to $k$ dimension. This elimination process is also called {\it Truncated SVD}, as described in~\cref{th:tsvd}.

\begin{theorem}[Truncated SVD and Matrix (Cayley) Multiplication~\citep{axler2024linear}]\label{th:tsvd}{
Given rank $r$, Truncated SVD for a linear transformation ${\bf T}$ is given by ${\bf T} {\bf v}_i  = s_i v_i {\bf w}_i$, $i=1,2,\ldots,r$. Suppose ${\bf v}_i \in \mathbb{R}^{m}$ and  ${\bf w}_i \in \mathbb{R}^{n}$,and denote $ {\bf V}_r = ({\bf v}_1, {\bf v}_2,\ldots,{\bf v}_r) \in \mathbb{R}^{m \times r}$ and $ {\bf W}_r = (s_1 v_1 {\bf w}_1, s_2 v_2 {\bf w}_2,\ldots, s_r v_r {\bf w}_r) \in \mathbb{R}^{n \times r}$, so ${\bf T} {\bf V}_r = {\bf W}_r \Rightarrow {\bf T} = {\bf W}_r {\bf V}^{\ast}_r$ is a matrix multiplication form of ${\bf T}$, where ${\bf V}^{\ast}_r$ is the conjugate transpose of ${\bf V}_r$. 

}
\end{theorem}

\begin{figure}[h!]
    \centering

    \begin{subfigure}[b]{\textwidth}
        \centering
        \begin{tikzpicture}

    \node[color=RoyalBlue] at (-0.8,0,-0.25) {$\mathbb{W} \subset \mathbb{R}^{n}$};
    \node[color=gray] at (1.7,0.8,0) {${\bf W}_r^{\top} $};
    \node[color=gray] at (5.3,0.8,0) {$({\bf V}_r^{\ast})^{\top} $};
    \node[color=Maroon] at (2.5,0,-0.25) {$\mathbb{W}_r \subset \mathbb{R}^{r}$};
    \node[color=OliveGreen] at (7.6,0,-0.25) {$\mathbb{V}^{\ast} \subset \mathbb{R}^{m}$};

    \draw[ultra thick,-stealth] (0,0,0) -- (0.5,0.5,0.5) node[anchor=south]{$\bf{w}$};
    \draw[line width=0.2mm, -stealth,RoyalBlue] (0,0,0) -- (1,0,0) node[anchor=north east]{};
    \draw[line width=0.2mm,-stealth,RoyalBlue] (0,0,0) -- (0,1,0) node[anchor=north west]{};
    \draw[line width=0.2mm,-stealth,RoyalBlue] (0,0,0) -- (0,0,1) node[anchor=east]{};

    \draw[ultra thick, -stealth, gray] (1.2,0.5,0) -- (2.2,0.5,0);

    \draw[line width=0.2mm, -stealth,Maroon] (3.7,0.2,0) -- (4.2,-0.25,-0.25) node[anchor=north east]{};
    \draw[line width=0.2mm,-stealth,Maroon] (3.7,0.2,0) -- (4,0.8,0) node[anchor=north west]{};
    \draw[line width=0.2mm,-stealth,Maroon] (3.7,0.2,0) -- (3.4,0,1) node[anchor=east]{};

    \draw[ultra thick,-stealth] (3.7,0.2,0) -- (3.4,0.6,0.5) node[anchor=south]{$\bf{r}$};

    \draw[ultra thick, -stealth, gray] (4.7,0.5,0) -- (5.7,0.5,0);

    \draw[line width=0.2mm, -stealth,OliveGreen] (7,0.2,0) -- (6.5,0.5,1) node[anchor=north east]{};
    \draw[line width=0.2mm,-stealth,OliveGreen] (7,0.2,0) -- (7,0.3,-1) node[anchor=north west]{};
    \draw[line width=0.2mm,-stealth,OliveGreen] (7,0.2,0) -- (7.5,0.2,1) node[anchor=east]{};

        \draw[ultra thick,-stealth] (7,0.2,0) -- (6.5,0.7,0.1) node[anchor=east]{${\bf v}$};
    
\end{tikzpicture}
        \caption{Matrix (Cayley) multiplication ${\bf W}_r {\bf V}_r^{\ast}: \mathbb{R}^{n} \rightarrow \mathbb{R}^{r} \rightarrow \mathbb{R}^{m}$ in~\cref{th:tsvd}, where ${\bf W}_r \in \mathbb{R}^{n \times r} $ and ${\bf V}_r^{\ast} \in \mathbb{R}^{r \times m} $}
        \label{fig:cayley}
\end{subfigure}

\begin{subfigure}[b]{\textwidth}
        \centering
        \input{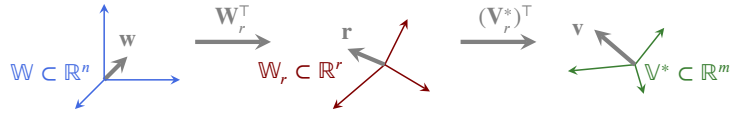}
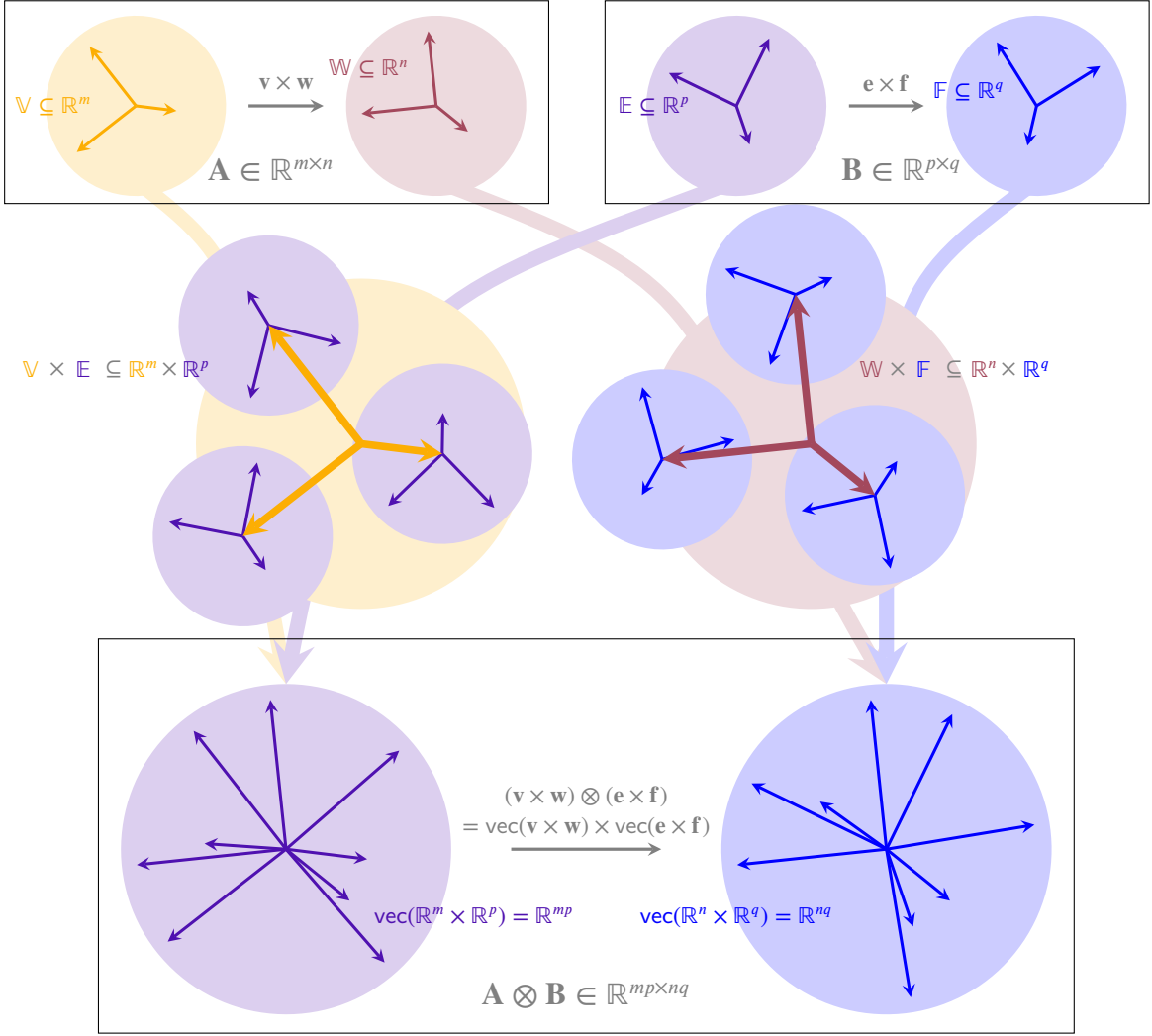
        \caption{Kronecker product ${\bf A}\otimes {\bf B}: \mathbb{R}^{m \times n} \rightarrow \mathbb{R}^{mp\times nq}$. The matrix ${\bf A}$ and ${\bf B}$ are composed according to~\cref{def:mat-comp}. That is, the first matrix, ${\bf A}$, is composed as ${\bf A}={\bf v} \times {\bf w}$, and the second matrix, ${\bf B}$ is composed as ${\bf B}={\bf e} \times {\bf f}$. The two vectors that compose matrix ${\bf A}$ serve as ``parent nodes'' and the two vectors that compose matrix ${\bf B}$ serve as ``child nodes''. The final results of Kronecker products ${\bf A}\otimes {\bf B}$ is the composition of all the ``child nodes''. Here, the four vectors ${\bf v} \in \mathbb{R}^m$, ${\bf w} \in \mathbb{R}^n$,  ${\bf e} \in \mathbb{R}^p$ and ${\bf f} \in \mathbb{R}^q$.
        }
        \label{fig:kron}
\end{subfigure}
\caption{Matrix (Cayley) multiplication and Kronecker product.}
\end{figure}

\subsubsection{Kronecker Product}

Until now, we have assumed that all the subspaces are equivalent in the composition of a matrix or tensor. However, there should be some cases where some subspaces work on top the another, or in other words, cases of hierarchies. The Kronecker product is such a case, where all the elements of the first operands are scalar multiplied by the second operand, as shown in~\cref{fig:kron}. For two matrices ${\bf A}\in \mathbb{R}^{m\times n}$ and ${\bf B} \in \mathbb{R}^{p \times q}$, is the Kronecker product of ${\bf A}$ and ${\bf B}$ is 
\begin{equation}\label{eq:kron}
    {\bf A} \otimes {\bf B} =
    \left[\begin{array}{ccc}
        a_{11}{B} & \cdots & a_{1n}{B} \\ 
        \vdots & \ddots & \vdots\\
        a_{m1}{B} & \cdots & a_{mn}{B} \\
\end{array}\right] 
\in \mathbb{R}^{mp \times nq}.
\end{equation}

From the matrix composition perspective (\cref{def:mat-comp}), suppose ${\bf A}={\bf v} \times {\bf w}$,  ${\bf B}={\bf e} \times {\bf f}$, then $(i,j)$th element of $A \otimes B$ is given by
\begin{equation}
(A \otimes B)_{i,j} = a_{\lceil i/p \rfloor, \lceil j/q \rfloor} b_{(i-1)\%{p}+1, (j-1)\%{q}+1 } = v_{\lceil i/p \rfloor} w_{\lceil j/q \rfloor} \Bigl( e_{(i-1)\%{p}+1} f_{(j-1)\%{q}+1} \Bigr),
\end{equation}
where $\%$ denotes the remainder when dividing. Moreover, its representation with four vectors ${\bf v \in \mathbb{V}}$, ${\bf w \in \mathbb{W}}$, ${\bf e \in \mathbb{E}}$ and ${\bf f \in \mathbb{F}}$ is
\begin{equation}
    {\bf A} \otimes {\bf B} = \operatorname{vec}({\bf v}\times {\bf e}) \times \operatorname{vec}({\bf w}\times {\bf f}).
\end{equation}

\cref{fig:kron} illustrates the hierarchy of the Kronecker product, where ${\bf e}$ and ${\bf f}$ can be seen as the replicated child nodes of ${\bf v}$ and ${\bf w}$, respectively. The final result is composed of flattened ``child nodes'', however, all ``child nodes'' are globally organized by ``parent nodes'' ${\bf v}$ and ${\bf w}$. 

\subsubsection{Block-diagonal Matrices}\label{sec:block-diagonal}

Block-diagonal matrices are blocks on the matrix diagonal while the rest elements are zeros. A remarkable example of such kind of matrices for model compression is butterfly matrix~\cite{Dao2022MonarchES,Dao2020KaleidoscopeAE,Dao2019LearningFA,de2018two}, which can represent a matrix ${\bf C}$ with~\cref{eq:map} as~\cref{eq:block}, where ${\bf C}_1 \in \mathbb{R}^{m\times m}$ and ${\bf C}_2 \in \mathbb{R}^{n\times n}$, ${\bf P}$ is a permutation matrix, and ${\bf D}$ is a butterfly factor~\cite{Dao2019LearningFA} to yield 
\begin{equation}\label{eq:block}
    {\bf C}^{\top}= \left[\begin{array}{cc}
        {\bf D}_1 & {\bf D}_2  \\ 
        {\bf D}_3 & {\bf D}_4  \\
\end{array}\right]\left[\begin{array}{cc}
        {\bf C}_1 & 0  \\ 
        0 & {\bf C}_2  \\
\end{array}\right]{\bf P}.\\
\end{equation}

According to~\cref{def:mat-comp}, $\left[\begin{array}{cc}
        {\bf C}_1 & 0  \\ 
        0 & {\bf C}_2  \\
\end{array}\right] = \left[\begin{array}{cc}
        {\bf C}_1 & 0  \\ 
        0 & 0  \\
\end{array}\right] + \left[\begin{array}{cc}
        0 & 0  \\ 
        0 & {\bf C}_2  \\
\end{array}\right] $ can be seen as the vector addition between two vectors ${\bf c}_1,  {\bf c}_2 \in {\mathbb{R}}^{{(m+n)}\times{(m+n)}}$. In~\cref{sec:params}, we show that ${\bf c}_1$ and ${\bf c}_2$ are composed by ${\bf a}_1 = (a_1, a_2, \ldots, a_m, \overbrace{0, \ldots, 0}^{m+2n})$ and ${\bf b}_1 = (\overbrace{0, \ldots, 0}^{m}, b_1, b_2, \ldots, b_m , \overbrace{0, \ldots, 0}^{2n})$,  ${\bf a}_2 = (\overbrace{0, \ldots, 0}^{2m}, a_{m+1}, a_{m+2}, \ldots, a_{m+n}, \overbrace{0, \ldots, 0}^{n})$ and ${\bf b}_2 = (\overbrace{0, \ldots, 0}^{2m+n}, b_{m+1}, b_{m+2}, \ldots, b_{m+n})$ in a unified vector space ${\mathbb V} \subseteq \mathbb{R}^{2m+2n}$. 
This process can be further interpreted as the composition in two different subspaces ${\mathbb V}_1 \subseteq \mathbb{R}^{m\times m}$ and ${\mathbb V}_2 \subseteq \mathbb{R}^{n \times n}$, and ${\mathbb V} = {\mathbb V}_1 \times {\mathbb V}_2$ (regarding coordinates), with ${\bf a}_1, {\bf b}_1 \in \mathbb{V}_1$ and  ${\bf a}_2, {\bf b}_2 \in \mathbb{V}_2$. 

There may be transformations in $\mathbb{V}_1$ and $\mathbb{V}_2$ separately, which implies that in hardware the two processes can run in parallel. Then, the vectors ${\bf a}_1$, ${\bf b}_1$ and ${\bf a}_2$, ${\bf b}_2$ in two subspaces $\mathbb{V}_1$ and $\mathbb{V}_2$ are merged together and go through a transformation defined by ${\bf D}$. 

\cref{eq:block} can be further expanded as
\begin{equation}
{\bf C}^{\top}= {
\left[\begin{array}{ccc|ccc}
        d_{1,1} &  &  & d_{2,1} &  &    \\ 
         & \ddots &  &  & \ddots &  \\ 
        &  & d_{1,m} & & & d_{2,n}   \\ \hline
        d_{3,1} &  &  & d_{4,1} &  &  \\
        & \ddots & &  & \ddots &   \\
        & & d_{3,m} &  & & d_{4,b}  \\ 
\end{array}\right]
}
{\left[\begin{array}{cccccc}
        a_1 b_1 & \ldots & a_m b_1 & \multicolumn{3}{c}{\multirow{3}{*}{0}}   \\ 
        \vdots & \ddots & \vdots &   \\ 
        a_1 b_m & \ldots & a_m b_m &   \\
        \multicolumn{3}{c}{\multirow{3}{*}{0}}& a_{m+1}b_{m+1} & \ldots & a_{m+n}b_{m+1}  \\
        & & & \vdots & \ddots & \vdots  \\
        & & & a_{m+1}b_{m+n} & \ldots & a_{m+n}b_{m+n}  \\ 
\end{array}\right]}
\in \mathbb{R}^{(n+m)\times (n+m)}.
\end{equation}\label{eq:block-detailed}

According to~\cite{Dao2019LearningFA}, the left matrix in~\cref{eq:block-detailed} can be recursively factorized into smaller diagonal matrices, which finally results in a set of $2\times 2$ diagonal matrices. These $2\times 2$ diagonal matrices can be seen as $1$-dimensional vectors in order-$1$ vector space, or equivalently, as scalars with index information in the parameter vector. The multiplication of such scalars and ${\bf C}_1$ and ${\bf C}_2$ can be seen as scaling in different subspaces. Herein, block-diagonal matrices can be geometrically interpreted as separate scalar scaling in smaller subspaces, and its following merge across these subspaces with concatenation. 

\subsection{Tensor Factorization}\label{sec:td}

Compared with matrix factorizations in~\cref{sec:matd}, tensor factorizations are a bit more complex (but fortunately not too much), since there are more than two subspaces, and every two subspaces have interactions like in~\cref{th:svd,th:tsvd}. All the interactions are not outside those in~\cref{tab:op}, which makes the various tensor factorization orchestrated in the same style. The three tensor factorization formats we discuss in this section are the CP, Tucker and Tensor-Train. We later summarize that their forms represent roughly vector addition in the same subspace, dimension adjust intra subspace, and subspace order expansion.

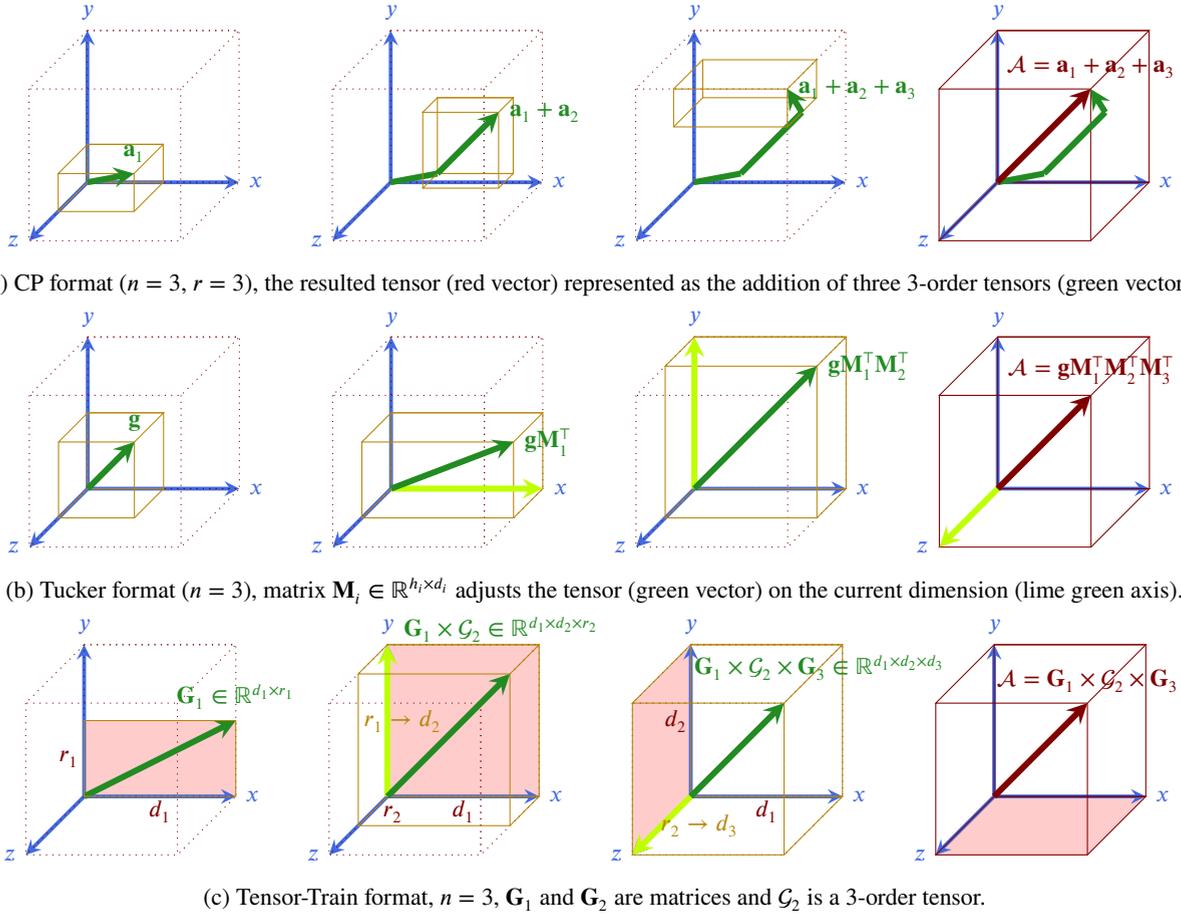
\begin{figure}[h!]
    \centering

    \begin{subfigure}[b]{\textwidth}
        \centering
            \begin{tikzpicture}

    \draw[line width=0.5mm, -stealth,RoyalBlue] (0,0,0) -- (2,0,0) node[anchor=west]{$x$};
    \draw[line width=0.5mm,-stealth,RoyalBlue] (0,0,0) -- (0,2,0) node[anchor=south]{$y$};
    \draw[line width=0.5mm,-stealth,RoyalBlue] (0,0,0) -- (0,0,2) node[anchor=east]{$z$};
    \draw[color=Maroon, dotted] (0,0,0) -- (2,0,0) -- (2,2,0) -- (0,2,0) -- cycle;
    \draw[color=Maroon, dotted] (0,0,0) -- (2,0,0) -- (2,0,2) -- (0,0,2) -- cycle;
    \draw[color=Maroon, dotted] (0,2,0) -- (2,2,0) -- (2,2,2) -- (0,2,2) -- cycle;
    \draw[color=Maroon, dotted] (0,2,2) -- (0,0,2);
    \draw[color=Maroon, dotted] (2,2,2) -- (2,0,2);

    \draw[color=DarkGoldenrod] (0,0,0) -- (1,0,0) -- (1,0.5,0) -- (0,0.5,0) -- cycle;
    \draw[color=DarkGoldenrod] (0,0,0) -- (1,0,0) -- (1,0,1) -- (0,0,1) -- cycle;
    \draw[color=DarkGoldenrod] (0,0.5,0) -- (1,0.5,0) -- (1,0.5,1) -- (0,0.5,1) -- cycle;
    \draw[color=DarkGoldenrod] (0,0.5,1) -- (0,0,1);
    \draw[color=DarkGoldenrod] (1,0.5,1) -- (1,0,1);
    
    \draw[line width=0.8mm,ForestGreen, -stealth] (0,0,0) -- (1,0.5,1)node[anchor=south]{${\bf a}_1$};

    \draw[line width=0.5mm, -stealth,RoyalBlue] (4,0,0) -- (6,0,0) node[anchor=west]{$x$};
    \draw[line width=0.5mm,-stealth,RoyalBlue] (4,0,0) -- (4,2,0) node[anchor=south]{$y$};
    \draw[line width=0.5mm,-stealth,RoyalBlue] (4,0,0) -- (4,0,2) node[anchor=east]{$z$};
    \draw[color=Maroon, dotted] (4,0,0) -- (6,0,0) -- (6,2,0) -- (4,2,0) -- cycle;
    \draw[color=Maroon, dotted] (4,0,0) -- (6,0,0) -- (6,0,2) -- (4,0,2) -- cycle;
    \draw[color=Maroon, dotted] (4,2,0) -- (6,2,0) -- (6,2,2) -- (4,2,2) -- cycle;
    \draw[color=Maroon, dotted] (4,2,2) -- (4,0,2);
    \draw[color=Maroon, dotted] (6,2,2) -- (6,0,2);
    \draw[color=Maroon, dotted] (6,2,2) -- (6,0,2);
    
    \draw[line width=0.8mm,ForestGreen] (4,0,0) -- (5,0.5,1);
    \draw[color=DarkGoldenrod] (5,0.5,1) -- (6,0.5,1) -- (6,1.5,1) -- (5,1.5,1) -- cycle;
    \draw[color=DarkGoldenrod] (5,0.5,1) -- (6,0.5,1) -- (6,0.5,1.5) -- (5,0.5,1.5) -- cycle;
    \draw[color=DarkGoldenrod] (5,1.5,1) -- (6,1.5,1) -- (6,1.5,1.5) -- (5,1.5,1.5) -- cycle;
    \draw[color=DarkGoldenrod] (5,1.5,1.5) -- (5,0.5,1.5);
    \draw[color=DarkGoldenrod] (6,1.5,1.5) -- (6,0.5,1.5);
    \draw[line width=0.8mm,ForestGreen, -stealth] (5,0.5,1) -- (6,1.5,1.5)node[anchor=west]{${\bf a}_1+{\bf a}_2$};

    \draw[line width=0.5mm, -stealth,RoyalBlue] (8,0,0) -- (10,0,0) node[anchor=west]{$x$};
    \draw[line width=0.5mm,-stealth,RoyalBlue] (8,0,0) -- (8,2,0) node[anchor=south]{$y$};
    \draw[line width=0.5mm,-stealth,RoyalBlue] (8,0,0) -- (8,0,2) node[anchor=east]{$z$};
    \draw[color=Maroon, dotted] (8,0,0) -- (10,0,0) -- (10,2,0) -- (8,2,0) -- cycle;
    \draw[color=Maroon, dotted] (8,0,0) -- (10,0,0) -- (10,0,2) -- (8,0,2) -- cycle;
    \draw[color=Maroon, dotted] (8,2,0) -- (10,2,0) -- (10,2,2) -- (8,2,2) -- cycle;
    \draw[color=Maroon, dotted] (8,2,2) -- (8,0,2);
    \draw[color=Maroon, dotted] (10,2,2) -- (10,0,2);
    \draw[line width=0.8mm,ForestGreen] (8,0,0) -- (9,0.5,1);
    \draw[line width=0.8mm,ForestGreen] (9,0.5,1) -- (10,1.5,1.5);
    \draw[color=DarkGoldenrod] (8.5,1.5,2) -- (10,1.5,2) -- (10,2,2) -- (8.5,2,2) -- cycle;
    \draw[color=DarkGoldenrod] (8.5,1.5,1) -- (10,1.5,1) -- (10,2,1) -- (8.5,2,1) -- cycle;
    \draw[color=DarkGoldenrod] (10,2,2) -- (10,2,1);
    \draw[color=DarkGoldenrod] (8.5,2,2) -- (8.5,2,1);
    \draw[color=DarkGoldenrod] (10,1.5,2) -- (10,1.5,1);
    \draw[color=DarkGoldenrod] (8.5,1.5,2) -- (8.5,1.5,1);
    \draw[line width=0.8mm,ForestGreen, -stealth] (10,1.5,1.5) -- (10,2,2)node[anchor=west ]{${\bf a}_1+{\bf a}_2+{\bf a}_3$};

    \draw[line width=0.5mm, -stealth,RoyalBlue] (12,0,0) -- (14,0,0) node[anchor=west]{$x$};
    \draw[line width=0.5mm,-stealth,RoyalBlue] (12,0,0) -- (12,2,0) node[anchor=south]{$y$};
    \draw[line width=0.5mm,-stealth,RoyalBlue] (12,0,0) -- (12,0,2) node[anchor=east]{$z$};
    \draw[color=Maroon] (12,0,0) -- (14,0,0) -- (14,2,0) -- (12,2,0) -- cycle;
    \draw[color=Maroon] (12,0,0) -- (14,0,0) -- (14,0,2) -- (12,0,2) -- cycle;
    \draw[color=Maroon] (12,2,0) -- (14,2,0) -- (14,2,2) -- (12,2,2) -- cycle;
    \draw[color=Maroon] (12,2,2) -- (12,0,2);
    \draw[color=Maroon] (14,2,2) -- (14,0,2);
    \draw[line width=0.8mm,ForestGreen] (12,0,0) -- (13,0.5,1);
    \draw[line width=0.8mm,ForestGreen] (13,0.5,1) -- (14,1.5,1.5);
    \draw[line width=0.8mm,ForestGreen, -stealth] (14,1.5,1.5) -- (14,2,2);
    \draw[line width=0.8mm,Maroon, -stealth] (12,0,0) -- (14,2,2)node[anchor=south]{$\mathcal{A}={\bf a}_1+{\bf a}_2+{\bf a}_3$};

\end{tikzpicture}
        \caption{CP format ($n=3$, $r=3$), the resulted tensor (red vector) represented as the addition of three $3$-order tensors (green vectors).}
        \label{fig:cp}
\end{subfigure}
\begin{subfigure}[b]{\textwidth}
        \centering
        \begin{tikzpicture}

    \draw[line width=0.5mm, -stealth,RoyalBlue] (0,0,0) -- (2,0,0) node[anchor=west]{$x$};
    \draw[line width=0.5mm,-stealth,RoyalBlue] (0,0,0) -- (0,2,0) node[anchor=south]{$y$};
    \draw[line width=0.5mm,-stealth,RoyalBlue] (0,0,0) -- (0,0,2) node[anchor=east]{$z$};
    \draw[color=Maroon, dotted] (0,0,0) -- (2,0,0) -- (2,2,0) -- (0,2,0) -- cycle;
    \draw[color=Maroon, dotted] (0,0,0) -- (2,0,0) -- (2,0,2) -- (0,0,2) -- cycle;
    \draw[color=Maroon, dotted] (0,2,0) -- (2,2,0) -- (2,2,2) -- (0,2,2) -- cycle;
    \draw[color=Maroon, dotted] (0,2,2) -- (0,0,2);
    \draw[color=Maroon, dotted] (2,2,2) -- (2,0,2);

    \draw[color=DarkGoldenrod] (0,0,0) -- (1,0,0) -- (1,1,0) -- (0,1,0) -- cycle;
    \draw[color=DarkGoldenrod] (0,0,0) -- (1,0,0) -- (1,0,1) -- (0,0,1) -- cycle;
    \draw[color=DarkGoldenrod] (0,1,0) -- (1,1,0) -- (1,1,1) -- (0,1,1) -- cycle;
    \draw[color=DarkGoldenrod] (0,1,1) -- (0,0,1);
    \draw[color=DarkGoldenrod] (1,1,1) -- (1,0,1);
    
    \draw[line width=0.8mm,ForestGreen, -stealth] (0,0,0) -- (1,1,1)node[anchor=south]{${\bf g}$};

    
    \draw[line width=0.5mm,-stealth,RoyalBlue] (4,0,0) -- (4,2,0) node[anchor=south]{$y$};
    \draw[line width=0.5mm,-stealth,RoyalBlue] (4,0,0) -- (4,0,2) node[anchor=east]{$z$};
    \draw[color=Maroon, dotted] (4,0,0) -- (6,0,0) -- (6,2,0) -- (4,2,0) -- cycle;
    \draw[color=Maroon, dotted] (4,0,0) -- (6,0,0) -- (6,0,2) -- (4,0,2) -- cycle;
    \draw[color=Maroon, dotted] (4,2,0) -- (6,2,0) -- (6,2,2) -- (4,2,2) -- cycle;
    
    \draw[color=Maroon, dotted] (4,2,2) -- (4,0,2);
    \draw[color=Maroon, dotted] (6,2,2) -- (6,0,2);
    \draw[color=Maroon, dotted] (6,2,2) -- (6,0,2);
    
    \draw[color=DarkGoldenrod] (4,0,0) -- (6,0,0) -- (6,1,0) -- (4,1,0) -- cycle;
    \draw[color=DarkGoldenrod] (4,0,0) -- (4,0,1) -- (6,0,1) -- (6,0,0) -- cycle;
    \draw[color=DarkGoldenrod] (4,1,0) -- (4,1,1) -- (6,1,1) -- (6,1,0) -- cycle;
    \draw[color=DarkGoldenrod] (4,1,1) -- (4,0,1);
    \draw[color=DarkGoldenrod] (6,1,1) -- (6,0,1);
    \draw[line width=0.8mm, -stealth,lime] (4,0,0) -- (6,0,0) node[anchor=west]{\textcolor{RoyalBlue}{$x$}};
    
    \draw[line width=0.8mm,ForestGreen, -stealth] (4,0,0) -- (6,1,1)node[anchor=west]{${\bf g}{\bf M}_1^{\top}$};

    
    \draw[line width=0.5mm, -stealth,RoyalBlue] (8,0,0) -- (10,0,0) node[anchor=west]{$x$};
    
    \draw[line width=0.5mm,-stealth,RoyalBlue] (8,0,0) -- (8,0,2) node[anchor=east]{$z$};
    \draw[color=Maroon, dotted] (8,0,0) -- (10,0,0) -- (10,2,0) -- (8,2,0) -- cycle;
    \draw[color=Maroon, dotted] (8,0,0) -- (10,0,0) -- (10,0,2) -- (8,0,2) -- cycle;
    \draw[color=Maroon, dotted] (8,2,0) -- (10,2,0) -- (10,2,2) -- (8,2,2) -- cycle;
    \draw[color=Maroon, dotted] (8,2,2) -- (8,0,2);
    \draw[color=Maroon, dotted] (10,2,2) -- (10,0,2);

        \draw[color=DarkGoldenrod] (8,0,0) -- (10,0,0) -- (10,2,0) -- (8,2,0) -- cycle;
    \draw[color=DarkGoldenrod] (8,0,0) -- (8,0,1) -- (10,0,1) -- (10,0,0) -- cycle;
    \draw[color=DarkGoldenrod] (8,2,0) -- (8,2,1) -- (10,2,1) -- (10,2,0) -- cycle;
    \draw[color=DarkGoldenrod] (8,2,1) -- (8,0,1);
    \draw[color=DarkGoldenrod] (10,2,1) -- (10,0,1);

    \draw[line width=0.8mm,-stealth,lime] (8,0,0) -- (8,2,0) node[anchor=south]{\textcolor{RoyalBlue}{$y$}};
    \draw[line width=0.8mm,ForestGreen, -stealth] (8,0,0) -- (10,2,1)node[anchor=west]{${\bf g}{\bf M}_1^{\top}{\bf M}_2^{\top}$};

    \draw[line width=0.5mm, -stealth,RoyalBlue] (12,0,0) -- (14,0,0) node[anchor=west]{$x$};
    \draw[line width=0.5mm,-stealth,RoyalBlue] (12,0,0) -- (12,2,0) node[anchor=south]{$y$};
    
    \draw[color=Maroon] (12,0,0) -- (14,0,0) -- (14,2,0) -- (12,2,0) -- cycle;
    \draw[color=Maroon] (12,0,0) -- (14,0,0) -- (14,0,2) -- (12,0,2) -- cycle;
    \draw[color=Maroon] (12,2,0) -- (14,2,0) -- (14,2,2) -- (12,2,2) -- cycle;
    \draw[color=Maroon] (12,2,2) -- (12,0,2);
    \draw[color=Maroon] (14,2,2) -- (14,0,2);
    \draw[line width=0.8mm,-stealth,lime] (12,0,0) -- (12,0,2) node[anchor=east]{\textcolor{RoyalBlue}{$z$}};
    \draw[line width=0.8mm,Maroon, -stealth] (12,0,0) -- (14,2,2)node[anchor=south]{$\mathcal{A}={\bf g}{\bf M}_1^{\top}{\bf M}_2^{\top}{\bf M}_3^{\top}$};

\end{tikzpicture}
        \caption{Tucker format ($n=3$), matrix ${\bf M}_i \in \mathbb{R}^{h_i \times d_i}$ adjusts the tensor (green vector) on the current dimension (lime green axis).}
        \label{fig:tucker}
\end{subfigure}

\begin{subfigure}[b]{\textwidth}
        \centering
        \begin{tikzpicture}

    \fill[red!20] (0,0,0) -- (0,1,0) -- (2,1,0) -- (2,0,0) -- cycle;
    \draw[line width=0.5mm, -stealth,RoyalBlue] (0,0,0) -- (2,0,0) node[anchor=west]{$x$};
    \draw[line width=0.5mm,-stealth,RoyalBlue] (0,0,0) -- (0,2,0) node[anchor=south]{$y$};
    \draw[line width=0.5mm,-stealth,RoyalBlue] (0,0,0) -- (0,0,2) node[anchor=east]{$z$};
    \draw[color=Maroon, dotted] (0,0,0) -- (2,0,0) -- (2,2,0) -- (0,2,0) -- cycle;
    \draw[color=Maroon, dotted] (0,0,0) -- (2,0,0) -- (2,0,2) -- (0,0,2) -- cycle;
    \draw[color=Maroon, dotted] (0,2,0) -- (2,2,0) -- (2,2,2) -- (0,2,2) -- cycle;
    \draw[color=Maroon, dotted] (0,2,2) -- (0,0,2);
    \draw[color=Maroon, dotted] (2,2,2) -- (2,0,2);

    \draw[line width=0.8mm,ForestGreen, -stealth] (0,0,0) -- (2,1,0)node[anchor=south]{${\bf G}_1 \in \mathbb{R}^{d_1 \times r_1}$};
    \draw[color=DarkGoldenrod] (0,0,0) -- (0,1,0) -- (2,1,0) -- (2,0,0) -- cycle;

    \node[color=Maroon] at (1,-0.2,0) {$d_1$};
    \node[color=Maroon] at (-0.2,0.5,0) {$r_1$};

    

    
    \fill[red!20] (4,0,0) -- (4,2,0) -- (6,2,0) -- (6,0,0) -- cycle;
    \draw[line width=0.5mm, -stealth,RoyalBlue] (4,0,0) -- (6,0,0) node[anchor=west]{$x$};
    \draw[line width=0.5mm,-stealth,RoyalBlue] (4,0,0) -- (4,2,0) node[anchor=south]{$y$};

    \draw[line width=0.5mm,-stealth,RoyalBlue] (4,0,0) -- (4,0,2) node[anchor=east]{$z$};
    
    \draw[color=Maroon, dotted] (4,0,0) -- (6,0,0) -- (6,2,0) -- (4,2,0) -- cycle;
    \draw[color=Maroon, dotted] (4,0,0) -- (6,0,0) -- (6,0,2) -- (4,0,2) -- cycle;
    \draw[color=Maroon, dotted] (4,2,0) -- (6,2,0) -- (6,2,2) -- (4,2,2) -- cycle;
    \draw[color=Maroon, dotted] (4,2,2) -- (4,0,2);
    \draw[color=Maroon, dotted] (6,2,2) -- (6,0,2);
    \draw[color=Maroon, dotted] (6,2,2) -- (6,0,2);

    \draw[color=DarkGoldenrod] (4,0,0) -- (6,0,0) -- (6,2,0) -- (4,2,0) -- cycle;
    \draw[line width=0.8mm,-stealth,lime] (4,0,0) -- (4,2,0) node[anchor=east]{};
    \draw[color=DarkGoldenrod] (4,0,0) -- (4,0,1) -- (6,0,1) -- (6,0,0) -- cycle;
    \draw[color=DarkGoldenrod] (4,2,0) -- (4,2,1) -- (6,2,1) -- (6,2,0) -- cycle;
    \draw[color=DarkGoldenrod] (4,2,1) -- (4,0,1);
    \draw[color=DarkGoldenrod] (6,2,1) -- (6,0,1);

    \node[color=Maroon] at (5,-0.2,0) {$d_1 $};
    \node[color=DarkGoldenrod] at (4.2,1,0) {$r_1 \rightarrow d_2 $};
    \node[color=Maroon] at (4.3,0,0.6) {$r_2 $};

     \draw[line width=0.8mm,ForestGreen, -stealth] (4,0,0) -- (6,2,1);
     \node[color=ForestGreen] at (5.5,2.2,0) {${\bf G}_1 \times \mathcal{G}_2 \in \mathbb{R}^{d_1 \times d_2 \times r_2}$};


    \fill[red!20] (8,2,0) -- (8,2,2) -- (8,0,2) -- (8,0,0) -- cycle;
    \draw[line width=0.5mm, -stealth,RoyalBlue] (8,0,0) -- (10,0,0) node[anchor=west]{$x$};
    \draw[line width=0.5mm,-stealth,RoyalBlue] (8,0,0) -- (8,2,0) node[anchor=south]{$y$};
    \draw[line width=0.5mm,-stealth,RoyalBlue] (8,0,0) -- (8,0,2) node[anchor=east]{$z$};
    \draw[color=Maroon, dotted] (8,0,0) -- (10,0,0) -- (10,2,0) -- (8,2,0) -- cycle;
    \draw[color=Maroon, dotted] (8,0,0) -- (10,0,0) -- (10,0,2) -- (8,0,2) -- cycle;
    \draw[color=Maroon, dotted] (8,2,0) -- (10,2,0) -- (10,2,2) -- (8,2,2) -- cycle;
    \draw[color=Maroon, dotted] (8,2,2) -- (8,0,2);
    \draw[color=Maroon, dotted] (10,2,2) -- (10,0,2);

    \draw[color=DarkGoldenrod] (8,0,0) -- (10,0,0) -- (10,2,0) -- (8,2,0) -- cycle;
    
    \draw[color=DarkGoldenrod] (8,0,0) -- (8,0,2) -- (10,0,2) -- (10,0,0) -- cycle;
    \draw[line width=0.8mm,-stealth,lime] (8,0,0) -- (8,0,2) node[anchor=east]{};
    \draw[color=DarkGoldenrod] (8,2,0) -- (8,2,2) -- (10,2,2) -- (10,2,0) -- cycle;
    
    \draw[color=DarkGoldenrod] (8,2,2) -- (8,0,2);
    \draw[color=DarkGoldenrod] (10,2,2) -- (10,0,2);

    \node[color=Maroon] at (9,-0.2,0) {$d_1 $};
    \node[color=Maroon] at (7.8,1,0) {$d_2 $};
    \node[color=DarkGoldenrod] at (8.5,0,1) {$r_2 \rightarrow d_3$};

    \draw[line width=0.8mm,ForestGreen, -stealth] (8,0,0) -- (10,2,2);
    \node[color=ForestGreen] at (9.7,1.7,0) {${\bf G}_1 \times \mathcal{G}_2 \times {\bf G}_3 \in \mathbb{R}^{d_1 \times d_2 \times d_3}$};

    \fill[red!20] (12,0,0) -- (12,0,2) -- (14,0,2) -- (14,0,0) -- cycle;
    \draw[line width=0.5mm, -stealth,RoyalBlue] (12,0,0) -- (14,0,0) node[anchor=west]{$x$};
    \draw[line width=0.5mm,-stealth,RoyalBlue] (12,0,0) -- (12,2,0) node[anchor=south]{$y$};
    \draw[line width=0.5mm,-stealth,RoyalBlue] (12,0,0) -- (12,0,2) node[anchor=east]{$z$};

    \draw[color=Maroon] (12,0,0) -- (14,0,0) -- (14,2,0) -- (12,2,0) -- cycle;
    
    \draw[color=Maroon] (12,0,0) -- (14,0,0) -- (14,0,2) -- (12,0,2) -- cycle;
    \draw[color=Maroon] (12,2,0) -- (14,2,0) -- (14,2,2) -- (12,2,2) -- cycle;
    
    \draw[color=Maroon] (12,2,2) -- (12,0,2);
    \draw[color=Maroon] (14,2,2) -- (14,0,2);
    \draw[line width=0.8mm,Maroon, -stealth] (12,0,0) -- (14,2,2)node[anchor=south]{$\mathcal{A} = {\bf G}_1 \times \mathcal{G}_2 \times {\bf G}_3$};

\end{tikzpicture}
        \caption{Tensor-Train format, $n=3$, ${\bf G}_1$ and ${\bf G}_2$ are matrices and $\mathcal{G}_2$ is a $3$-order tensor.}
        \label{fig:tt}
\end{subfigure}
\caption{Geometric interpretation of typical tensor decomposition formats, taking a $3$-order tensor as an example. Their vector forms are~\cref{eq:cp-vec,eq:tucker-vec}.
The \textcolor{DarkGoldenrod}{golden} cubes are the current constructed subsapces, the \textcolor{lime}{light green} vectors are the subspace expansion represented by shared order (e.g. $\mathcal{A}^{(i)} \cap {\bf M}_i$ in Tucker format, $\mathcal{A}^{(i)} \cap {\bf G}_i$ and $\mathcal{A}^{(i)} \cap \mathcal{G}_i$ in Tensor-Train format), the \textcolor{ForestGreen}{dark green} vectors are the current resulting vectors, inside the current constructed subspace. The \textcolor{red}{red} cube is the final resulted tensor, and \textcolor{red!20}{pink} planes represent resized dimension.}
\end{figure}

\subsubsection{CANDE-COMP/PARAFAC (CP) Format}
The CP format represents a tensor with a sum of tensor product of vectors~\cite{kolda2009tensor}. For a order-$n$ tensor, $\mathcal{A} \in \mathbb{R}^{d_1 \times d_2 \times \cdots \times d_n}$, we have $\mathcal{A} = \sum^{r}_{i=1} {\bf a}_i, \quad {\bf a}_i =  \overbrace{ {\bf v}_i \otimes {\bf u}_i \otimes \cdots \otimes {\bf w}_i }^{n \text{ vectors}},
$, 
where $r$ is the tensor rank and ${\bf v}_i \in \mathbb{R}^{d_1}$,   ${\bf u}_i \in \mathbb{R}^{d_2}$, $\ldots$, ${\bf w}_i \in \mathbb{R}^{d_n}$. 
We can easily convert this tensor format to a vector in $\mathbb{R}^{d_1 \times d_2 \times \cdots \times d_n}$ as
\begin{equation}\label{eq:cp-vec}
    {\bf a} = \sum^{r}_{i=1} {\bf a}^{(i)}, \qquad {\bf a}^{(i)}= {\bf a}^{(i)}_1 \times {\bf a}^{(i)}_2 \times \cdots \times {\bf a}^{(i)}_n.
\end{equation}

Such a format is an addition form of $r$ $n$-order tensors. All these order-$n$ tensors (represented as dark green individual vectors in~\cref{fig:cp}), are in the same subspace, as well as the resulting tensor $\mathcal{A}$ (represented as the final red vector in~\cref{fig:cp}). 

\subsubsection{Tucker Format}
For a order-$n$ tensor $\mathcal{A} \in \mathbb{R}^{d_1 \times d_2 \times \cdots \times d_n}$, its Tucker format is $\mathcal{A} = \mathcal{G} \times_1 {\bf M}_1 \times_2 \cdots \times_n {\bf M}_n$, 
where $\mathcal{G} \in \mathbb{R}^{l_1 \times l_2 \times \cdots \times l_n}$ is a $n$-order tensor and matrix ${\bf M}_i \in \mathbb{R}^{h_i \times d_i}$ ($i=1,2,\ldots, n$). 
Matrices ${\bf M}_i$ can be seen as a general linear transformation, ${\bf M}_i: \mathbb{R}^{h_i} \rightarrow \mathbb{R}^{d_i}$, which results in a transformation on the whole tensor $\mathcal{G} \times_1 {\bf M}_1 \times_2 \cdots \times_i {\bf M}_i: \mathbb{R}^{d_1 \times \cdots \times d_{i-1}  \times \textcolor{cyan}{h_i} \times h_{i+1} \times \cdots \times h_n} \rightarrow \mathbb{R}^{d_1 \times \cdots \times d_{i-1}  \times \textcolor{cyan}{d_i} \times h_{i+1} \times \cdots \times h_n}$. 

Suppose $({\bf v}_1, {\bf v}_2, \ldots, {\bf v}_n)$ is a basis of vector space $\mathbb{G} \in \mathbb{R}^{h_1 \times h_2 \times \cdots \times h_n}$, the vector form of Tucker format is 
\begin{equation}\label{eq:tucker-vec}
    {\bf a} = {\bf g}\times {\bf m}_1 \times {\bf m}_2 \times \cdots \times {\bf m}_n, \qquad {\bf m}_i \in \mathbb{M}_i \subseteq \mathbb{R}^{h_i \times d_i}, \quad
    \mathbb{G} \cap \mathbb{M}_i  = {\bf v}_i \in \mathbb{R}^{\max(h_i, d_i)}
\end{equation}
where ${\bf g} \in \mathbb{G}$.
As shown in~\cref{fig:tucker}, such a process represents changing the dimension range of each order of the subspace $\mathbb{G}$ via a tensor contraction mentioned in~\cref{eq:prod-basis}.  Geometrically, the product between the current tensor ${\bf g}\times {\bf m}_1 \times \cdots \times {\bf m}_{i-1}$ and ${\bf m}_i$ enlarges or shrinks the current dimension (highlighted on axis with lime green), reflecting the change of the current subspace (denoted as golden cubes in ~\cref{fig:tucker}). 

\subsubsection{Tensor-Train Format}
The Tensor-Train (TT) format~\citep{oseledets2011tensor} represents order-$n$ tensor $\mathcal{A} \in \mathbb{R}^{d_1 \times d_2 \times \cdots \times d_n}$ as $\mathcal{A} = {\bf G}_1 \times \mathcal{G}_2  \times^1_3 \cdots \times^1_3 \mathcal{G}_{N-1} \times {\bf G}_n$,
where $\mathcal{G}_i \in \mathbb{R}^{r_{i-1} \times d_i \times r_i}$, with the TT-rank set $\{ r_1, \ldots, r_{n-1}\}$, ${\bf G}_1 \in \mathbb{R}^{d_1 \times r_1}$ and ${\bf G}_n \in \mathbb{R}^{r_{n-1} \times d_{n}}$. The vector form of the Tensor-Train format is
\begin{equation}
    {\bf a} = {\bf m}_0 \times {\bf m}_1 \times \cdots {\bf m}_{n-1} \times {\bf m}_{n},\quad {\bf m}_i \in {\mathbb{M}_i},
\end{equation}
where $ \mathbb{M}_i \subset \{\mathbb{R}^{r_{i-1} \times d_i \times r_i}\big| i=2,3,\ldots, n-1\} \cup \{ \mathbb{R}^{d_i \times r_i} \big| i=1\} \cup \{ \mathbb{R}^{r_{i-1} \times d_i} \big| i=n \}$.

The vector form of Tensor-Train is quite simple, and the only point is that there is a dimension exceptions for ${\bf m}_0$ and ${\bf m}_n$ compared with other ${\bf m}_i$. Given that we have considered the tensor product and tensor contraction as identical in~\cref{eq:prod-basis}, where the contraction is a case where two subspaces share a dimension and ``absorb'' that dimension. In this sense, every product ${\bf m}_{i-1} \times {\bf m}_{i}$ means ``enlarging or shrinking the shared dimension between ${\bf m}_{i-1}$ and ${\bf{m}}_i$, which is similar to the Tucker format, and then additionally ``extend the new order introduced by ${\bf m}_i$'' when ${\bf m}_i$ is three-dimensional.

~\cref{fig:tt} illustrates the above geometric transformation, maintaining the matrix and tensor notations to make it easier for the reader to transit from the classical tensor version to our proposed vector version.
The first subspace is defined by ${\bf G}_1$, which is a matrix so that it settles down a 2-dimensional plane of size $d_1 \times r_1$. The product of ${\bf G}_1 \times \mathcal{G}_1$ is a subspace extension from 2-dimension to 3-dimension, along with the common dimension between ${\bf G}_1$ and $\mathcal{G}_1$ ($r_1$, highlighted with lime green $y$-axis). Similarly, the next product along with the common dimension between ${\bf G}_1 \times \mathcal{G}_1$ and ${\bf G}_3$, but it does not add extra dimension since ${\bf G}_3$ is a matrix which only scales the existing dimension.

\section{Our Taxonomy of Recent Research}\label{sec:literature}
We have addressed all the necessary issues regarding the terminologies of matrix/tensor factorization for model compression in~\cref{sec:application,sec:decompose}. Now we analyse relevant literature based on our proposed taxonomy. The core idea of our analysis is to \textbf{taking every matrix/tensor as geometric transformations}. Such analysis typically focuses on two questions:
\begin{enumerate}
    \item How many and what subspaces are involved?
    \item What geometric transformations (operations) occurs among these subspaces?
\end{enumerate}

In light of these two questions, we first revisit recent popular transformer models~\cref{sec:llm} with high-level geometric intuition, then interpret and compare matrix-based and tensor-based methodologies with our proposed taxonomy in~\cref{sec:compare}.

\subsection{GPT-2, Llama-3.1, Gemma-2 and Mistral Series Models}\label{sec:llm}

\begin{figure}
    \centering
    \includegraphics[width=\linewidth]{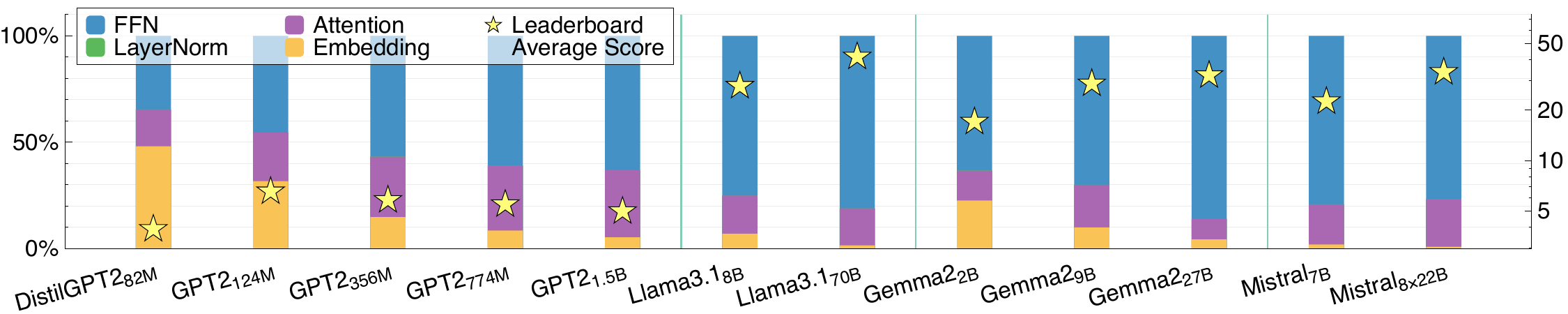}
    \caption{Parameter counts of different layer types of GPT2, Llama3.1, Gemma2 and Mistral series models, and their average benchmark scores on Huggingface Open LLM Leaderboard~\cite{open-llm-leaderboard-v2} (accessed on Sept 10, 2024). The y-axis on the left side indicates the parameter proportion, and that on the right side indicates the benchmark score. The benchmark scores represent the abilities of the LLMs to solve a set of complex tasks (e.g. reasoning, math problem-solving and QA). The higher the scores are, the better model performance on these tasks.}
    \label{fig:ratio-layers}
\end{figure}

The transformer structure underpins large language models (LLMs). We consider several popular open-sourced LLM series models (i.e. GPT2, Llama3.1, Gemma2 and Mistral) with available benchmark evaluation on~\cite{open-llm-leaderboard-v2}. These scores are the highest average values among different model instances (e.g. models fine-tuned on different datasets), with the listed model as the base model. We illustrate such scores and the parameter portion of different layer types of these base models in~\cref{fig:ratio-layers}, and explain the results in~\cref{fig:lm-mc,sec:lm-module}.

From~\cref{fig:ratio-layers}, we can observe that for the same series models, with the parameters of the whole model increasing, the ratio of the attention layers roughly remains the same (or slightly increases like for Gemma-2 and Mistral), while that of MLP increases. The ratio of embedding layers decreases when the total parameter counts increase, though, during our ratio calculation, the embedding layer parameter counts actually increase. Regarding the benchmark scores, we can observe that except for GPT-2 series models, the other three series models have increasing benchmark scores when the MLP parameter ratio account increases. Though the increasing total parameter counts might also be the reason for the increasing scores, we can explain the model performance with the different geometric transformations that different layer types have, which we discussed in~\cref{sec:lm-module}.

Recall that the attention layers map the input features into the value subspace $\mathbb{V}$, which is of a similar dimension as the query space $\mathbb{Q}$. However, the FFN in these LLMs typically maps the input features into a higher dimensional latent space first, and then to a lower-dimensional subspace, whose dimension is similar to the outputs of attention layers. The sequence of attention layers and MLP are shown in the right part of~\cref{fig:lm-mc}. 

\paragraph{For smaller LLMs like GPT-2, the model capability to memorize the distribution of data and more complex tasks (e.g. reasoning) does not diminish when the MLPs parameter ratios decrease.}   
From the discussion in~\cref{sec:subspaces} we know that all kinds of matrix multiplication (the same function that the dense layers in ~\cref{fig:ratio-layers}) can be seen as projections among different subspaces of the entire parameter space, thus the increasing ratio of the dense layers along with model size indicates the more and larger-scaled linear projections in GPT-2 series models. In practice, the model capability of ``memorization'' (i.e. tasks relevant to memorizing data like language modelling, summarization and translation) is improved with the increasing model size~\cite{radford2019language}. This improvement can be explained as the more and larger-scaled projections help to reduce the approximation error between the predictions and actual data distribution (i.e. ${\bf y}$ and $\hat{{\bf y}}$ in~\cref{fig:lm-mc}). Such an explanation is consistent with~\citep[Result 5]{AllenZhu2024PhysicsOL}, which reports that for GPT-2 models, reducing or even eliminating all the dense layers did not affect its model capacity for memorizing data. However, for more complex tasks like reasoning, these larger-scale projections might introduce more noise, as the HF scores decrease when the MLP ratios increase in~\cref{fig:ratio-layers}.      

\paragraph{For larger LLMs like Llama-3.1, Gemma-2 and Mistral, the model capability of more complex tasks (e.g. reasoning) diminishes when the MLPs parameter ratios decreases.} 
Unlike GPT-2, we didn't find the performance reports of tasks like language modeling of Llama-3.1, Gemma-2 and Mistral models, thus we cannot discuss the memorization capability of this models. However, we can explain the model performance of more complex tasks with those shown in~\cref{fig:ratio-layers}. The increase in attention layers can be explained as the model can memorize finer-grained data distribution, and the rise of FFN layers can be explained as the model's ability to deal with higher level, more abstract and implicit concepts and logic in the data, which might exist in the higher-dimension linear layers. Thus, for the benchmark scores, the model performance improves when the MLP ratio rises (sometimes also with increasing attention layers).   
\subsection{Comparison between Matrix and Tensor based Methodologies}\label{sec:compare}

We list the recent research works that use matrix factorization or tensor factorization to compress language models in~\cref{tab:compare}, these are mainly generative language models with a transformer structure (except LSTMs in~\cref{tab:compare}). The column ``Geometric Operations'' roughly corresponds to those in~\cref{sec:decompose}, and ``\#Subspaces'' means the number of subspaces per layer (the cross-layer exceptions are~\cite{NEURIPS2019_dc960c46,li-etal-2022-hypoformer}, where we take the considered layers as a whole).

\paragraph{Factual Analysis of the Comparison.}
In~\cref{tab:compare}, only two investigated works compressed all core modules (i.e. embedding, attention and feed-forward layers)~\cite{edalati2022kronecker,tahaei2021kroneckerbert}, this is because the forward passes of these three core modules are different, as we discussed in~\cref{sec:lm-module} and in practical implementation, this difference may introduce new problems. Most of the investigated works did not consider fidelity as their compression goal, though fidelity is calculated without actual language tasks and thus easier to obtain than accuracy in practice. This phenomenon indicates that most works believe fidelity is not a shortcut to accuracy and should be considered differently. Note that there are some works not have an optimization part in their methodologies at all~\cite{liu2021enabling,NEURIPS2019_dc960c46,li-etal-2022-hypoformer,liu2022tuformer}, mainly in the research work with tensor factorization. This is probably because the fidelity metrics for tensors (e.g. Frobenius norm) are less indicative of model accuracy than these for matrices, especially when the compared methodologies have different tensor orders. A more valid way to obtain the model accuracy is just selecting some hyperparameters (e.g. tensor ranks), running the compressed models, and evaluating the compressed model predictions.

\begin{landscape}
\begin{table}[]
\begin{tabular}{c|c|c|c|c|c|c|c|c|c}
\toprule

\multicolumn{1}{c|}{\multirow{2}{*}{\textbf{Work}}} & \multirow{2}{*}{\textbf{Base Model}} & \multicolumn{3}{c|}{\textbf{Layer Types}}                                                   & \multirow{2}{*}{\textbf{\#Subspaces}} & \multirow{2}{*}{\textbf{Geometric Operations}} & \multicolumn{2}{c|}{\textbf{Optimization Goal}} & \multirow{2}{*}{\begin{tabular}{cc}
     \textbf{Experimental Parameter}  \\
     \textbf{Compression Ratio}
\end{tabular}}  \\ \cline{3-5} \cline{8-9}
\multicolumn{1}{c|}{}                               &                                      & \multicolumn{1}{l|}{\textbf{Emb}} & \multicolumn{1}{l|}{\textbf{Attn}} & \textbf{FFN} &                                     &                                         & \multicolumn{1}{l|}{Fidelity} &    \multicolumn{1}{l|}{Accuracy} &                                 \\ \midrule
\multirow{2}{*}{\cite{Dao2022MonarchES}} & BERT &  & \multirow{2}{*}{$\surd$}  & \multirow{2}{*}{$\surd$}  & \multirow{2}{*}{3,4,5} & \multirow{2}{*}{\begin{tabular}{cc}
    generate $1$-order  \\
    subspaces, projection \end{tabular}} & \multirow{2}{*}{$\surd$} & & $0.98 \times$ (base)/ $1.33 \times$ (large)    \\ \cline{2-2} \cline{10-10}
       & GPT-2  &  &   &   &  & & &   &   $0.72 \times$ (small)/ $1.15 \times$ (medium)  \\ \hline
\cite{edalati2022kronecker}        & GPT-2 & $\surd$  & $\surd$  & $\surd$  & 3 & \multirow{2}{*}{\begin{tabular}{c} shrink subspace \\ dimensions \end{tabular} } &\multirow{2}{*}{$\surd$} &\multirow{2}{*}{$\surd$} &  $0.49 \times$ (small)   \\ \cline{1-6} \cline{10-10}
\cite{tahaei2021kroneckerbert}   &  BERT         &  $\surd$  &  $\surd$ & $\surd$ & 4  &   &  & &   $6.6\times - 18.3 \times$ (base)   \\ \hline
\cite{chen2018groupreduce}              & 2-layer LSTM  & $\surd$ & &   &  \multirow{6}{*}{2} &  \multirow{4}{*}{\begin{tabular}{c} split one subspace into two  \\ lower-dimensional subspaces \end{tabular}} & \multirow{2}{*}{$\surd$} & \multirow{2}{*}{$\surd$} &  $4\times - 9 \times$ (only Emb)   \\ \cline{1-5} \cline{10-10}
\cite{chen2021drone}              &  BERT  &  & $\surd$ &  $\surd$ &  &   & & &  {Not mentioned}   \\ \cline{1-5} \cline{8-10}
\cite{Lan2019ALBERTAL}              &  BERT  & $\surd$ & &   &  &   &   & &  $0.24\times$ (base) \\ \cline{1-5} \cline{8-10}
\cite{ben-noach-goldberg-2020-compressing}              &  BERT  &  &$\surd$  & $\surd$   &  &   &   & $\surd$ & $0.69\times$ (base) \\ \cline{1-5} \cline{7-10}
\cite{li2023losparse}              & \begin{tabular}{c}BERT, BART \\ DeBERTaV3\end{tabular}  &  &$\surd$ & $\surd$  &  & \begin{tabular}{c} split subspaces, use sparsity to \\reduce subspaces' dimensions \end{tabular}  &   & $\surd$ & $1\times - 19 \times$ \\ \cline{1-5} \cline{7-10}
\cite{wang2022exploring}              &  BERT, GPT-3& $\surd$ &$\surd$ & $\surd$  &  &  reallocate subspace dimensions  &   & & $0.3\times - 162.1\times$ \\ \hline
\cite{chekalina2023efficient}              &  GPT-2  &  &  & $\surd$  & 8 & \multirow{4}{*}{increase number of subspaces } &   & $\surd$  & $0.67\times$ (small) \\ \cline{1-6} \cline{8-10}
\cite{hrinchuk2019tensorize}              & \begin{tabular}{c} LSTM, \\ Transformers \end{tabular}   & $\surd$ &  &   & 6,8,12 &   &   & $\surd$ & \begin{tabular}{c} $7.88 \times$ (LSTM)/ \\ $2 \times$ (Transformers) \end{tabular} \\  \cline{1-6} \cline{8-10}
\cite{liu2021enabling}              & BERT   &  &$\surd$ & $\surd$  & 2,3,5,7 &   &  & & $0.52 \times - 0.64\times$ \\ \hline
\cite{NEURIPS2019_dc960c46}              &  Transformer  &   & $\surd$ &   & 3 & \multirow{2}{*}{change the projection sequence}  &   & & $1\times-2\times$ \\ \cline{1-6} \cline{8-10}
\cite{li-etal-2022-hypoformer}              &  Transformer  &  & $\surd$ & $\surd$  & 3 &   &   & & $2.6\times-4.5\times$ \\ \hline
\cite{liu2022tuformer}              &  Transformer  & $\surd$ & &   & 3 & reorganize linear transformations  &  & &   $0.24\times$ \\ \hline
\cite{qiucompute}              &  GPT-2  & & $\surd$ & $\surd$  & 3-20 & generate more 1-order subspaces   &  $\surd$ &  & Not mentioned \\
\bottomrule
\end{tabular}\caption{Interpretation of current literature interpreted under our proposed taxonomy. Since this paper focuses on model compression, we only extract the experimental model compression results claimed in the papers, with the compression ratio as defined in~\cref{def:mc}. The additional information after the compression ratio, for example, ``small'', ``large'', etc. reflects the different versions of the base models. The definition of two optimization goals,  fidelity and accuracy are given in~\cref{sec:lmc}. ``Not mentioned'' in the last column means the focus of the investigated work is not model compression (e.g. acceleration or reduced computations), but intrinsically reduces the model size at the same time. All the experimental results are comparable to language task performance in the uncompressed models.}\label{tab:compare}
\end{table}
\end{landscape}

\paragraph{Geometric View of the Comparison and Current Research Gap.} A straightforward idea of compressing neural networks is to directly break weight matrices into smaller matrices or tensors, which is equivalent to splitting high-dimensional subspaces into lower-dimensional subspaces. This idea and its variants can cover all aspects in~\cref{tab:compare}, except for those changing the subspace projection sequences across layers~\cite{NEURIPS2019_dc960c46,li-etal-2022-hypoformer}. From the viewpoint of subspaces on~\cref{tab:compare}, there are three obvious issues:
\begin{enumerate}
    \item What are the criteria of constructing subspaces according to a given neural network; 
    \item How to identify the less important subspaces, merge such subspaces with others, or even discard them;
    \item How to reallocate orders and dimensions in a collection of subspaces?
\end{enumerate}

The three questions are rather abstract and should be implemented with specific factorization methodologies like those in~\cref{sec:decompose}, as well as the actual forward pass of the considered neural networks. Furthermore, given that the target of model compression is to maintain comparable model performance with fewer parameters, it should be worthwhile to investigate the application requirements with the above three questions. For example, if the application requires the model to memorize more diverse data (i.e. more lower-dimensional subspaces in attention layers) or to have better reasoning ability (i.e. a few higher-dimensional subspaces in feed-forward layers).

\section{Conclusion}\label{sec:conclusion}
This paper has proposed a unified taxonomy related to the interpretation of the generative model compression, based on matrix and tensor factorizations. 
To bridge generative language model research and matrix/tensor factorization, and also bridge the matrix and tensor research, we have reduced the term subspace in geometric algebra as our pivot point to build these two connections.
To ease the understanding of the relevant algebra structure, we reduced and simplified the mathematical concepts as much as possible. For example, we have induced that the direct product, tensor product, Cartesian product and tensor contraction are the different sides of the same process - constructing a new vector space from other subspaces to avoid confusion caused by too many terms. 
We have finally provided the interpretation of the existing literature in the light of our subspace-oriented taxonomy and pointed out a roid when it comes to combining the required model ability (i.e. expressivity and capability) and the exact vector space implementations (i.e. certain types of matrix or tensor factorization). 

Since our focus has been on a unified taxonomy, we have not covered as vast amounts of literature as possible, and the unique details of each investigated work have not been fully discussed either. Also, some subspace-related insights are not easy to align with specific matrix/tensor factorization methods. Both are worthwhile for further investigation.

\bibliographystyle{unsrt}

\bibliography{ref}

\end{document}